\documentclass{article}

\usepackage[main, final]{neurips_2026}

\usepackage{graphicx}
\usepackage{makecell}
\usepackage{amssymb}
\usepackage{amsmath}
\usepackage{multirow}
\usepackage{tabularx}
\usepackage{subcaption}

\usepackage[utf8]{inputenc} 
\usepackage[T1]{fontenc}    
\usepackage{hyperref}       
\usepackage{url}            
\usepackage{booktabs}       
\usepackage{amsfonts}       
\usepackage{nicefrac}       
\usepackage{microtype}      
\usepackage{xcolor}         

\usepackage{cleveref}

\title{3D-VITAL: Visibility-aware Identity Training from Artificial Liftings of 3D Reconstruction model for Cross-View Object Re-Identification}

\author{%
    William Grolleau$^{1,2}$ \quad Astrid Sabourin$^{1}$ \quad Guillaume Lapouge$^{1}$ \quad Catherine Achard$^{2}$ \\
    $^{1}$Universit\'e Paris-Saclay, CEA, List, F-91120 Palaiseau, France \\
    $^{2}$Sorbonne University, CNRS, Institute of Intelligent Systems and Robotics (ISIR), Paris, France \\
    \texttt{\{william.grolleau,guillaume.lapouge\}@cea.fr}
}

\begin{document}
    \maketitle
    \definecolor{gain}{RGB}{34, 139, 34}
    \definecolor{loss}{RGB}{205, 0, 0}

    \newcommand{\resPos}[2]{\makecell{#1 ({\color{gain}+#2})}}
    \newcommand{\resNeg}[2]{\makecell{#1 ({\color{loss}-#2})}}

    \newcommand{\resPosBold}[2]{\makecell{\textbf{#1} ({\color{gain}\textbf{+#2}})}}
    \newcommand{\resNegBold}[2]{\makecell{\textbf{#1} ({\color{loss}\textbf{-#2}})}}

    \newcommand{\resSplitDown}[2]{ \makecell{#1\\\color{gain}{+#2}}}
    \newcommand{\resSplitDownBold}[2]{ \makecell{\textbf{#1}\\\color{gain}{\textbf{+#2}}}}

    \begin{abstract}
        In this paper, we address the Cross-View Object Re-Identification (ReID) task, aiming to match identities across drastically different viewpoints.
        We focus specifically on Single-View Aerial-Ground ReID (SV AG-ReID), where models train on one viewpoint and must generalize to another at inference.
        To bridge the viewpoint gap of Cross-View ReID, related work employs generative models to synthesize novel views from source images as additional training data.
        We leverage recent advances in large-scale single-view 3D mesh reconstruction to lift source images into textured meshes and to render them from new viewpoints.
        However, this pipeline produces unreliable data in two distinct ways, and addressing both is vital for synthetic data to serve as an effective training signal.
        First, the reconstruction may fail and produce geometrically incorrect meshes, resulting in severe corruption in rendered novel views.
        Second, even when the geometry is correct, the rendered novel views contain hallucinated textures in regions not visible in the source image, which may degrade training if used naively.
        We address both shortcomings by exploiting a signal that prior generative methods do not naturally expose.
        Indeed, the reconstructed mesh carries a per-pixel partition between regions visible from the source view and regions that the reconstruction model had to hallucinate, available as a deterministic byproduct of rendering.
        We introduce this per-pixel visibility map and exploit it in a two-stage design.
        Cross-View Consistency Filtering (CVCF) discards geometrically incorrect meshes by ensuring multi-view consistency between source and novel views over visible patches.
        Visibility-Aware Attention Supervision (VAAS) leverages the visibility map to guide the model's attention toward reliable regions of the rendered view during training.
        Combined, these two components enable our method, 3D-VITAL, to better leverage the synthetic signal from single-view 3D mesh reconstruction, benefiting from wider viewpoint augmentation.
        3D-VITAL sets a new state-of-the-art on three SV AG-ReID benchmarks (AG-ReID, AG-ReID.v2, MOO).
        Code and data will be released at \url{https://github.com/TurtleSmoke/3D-VITAL}.
    \end{abstract}

    \section{Introduction}

    Object Re-Identification (ReID) is a pattern recognition problem aiming to match the same object across non-overlapping cameras.
    This matching is especially challenging in Aerial-Ground ReID (AG-ReID), an emerging setting driven by the deployment of Unmanned Aerial Vehicles (UAVs).
    Here, the gap between top-down aerial views and frontal ground views causes severe geometric distortions, self-occlusions, and pose variations~\cite{zhang2020person, nguyen2023aerial, nguyen2024ag, shang2025aerial, Zhang_2024_CVPR, wang2025secap}.
    However, annotating data covering both viewpoints is costly, which has motivated the Single-View AG-ReID (SV AG-ReID) protocol~\cite{grolleau20263dlens3dliftingbasedelevated}, in which only one viewpoint is available during training and the model must generalize across unseen viewpoints at inference time.

    \begin{figure}[tbp]
        \centering
        \includegraphics[width=0.85\linewidth]{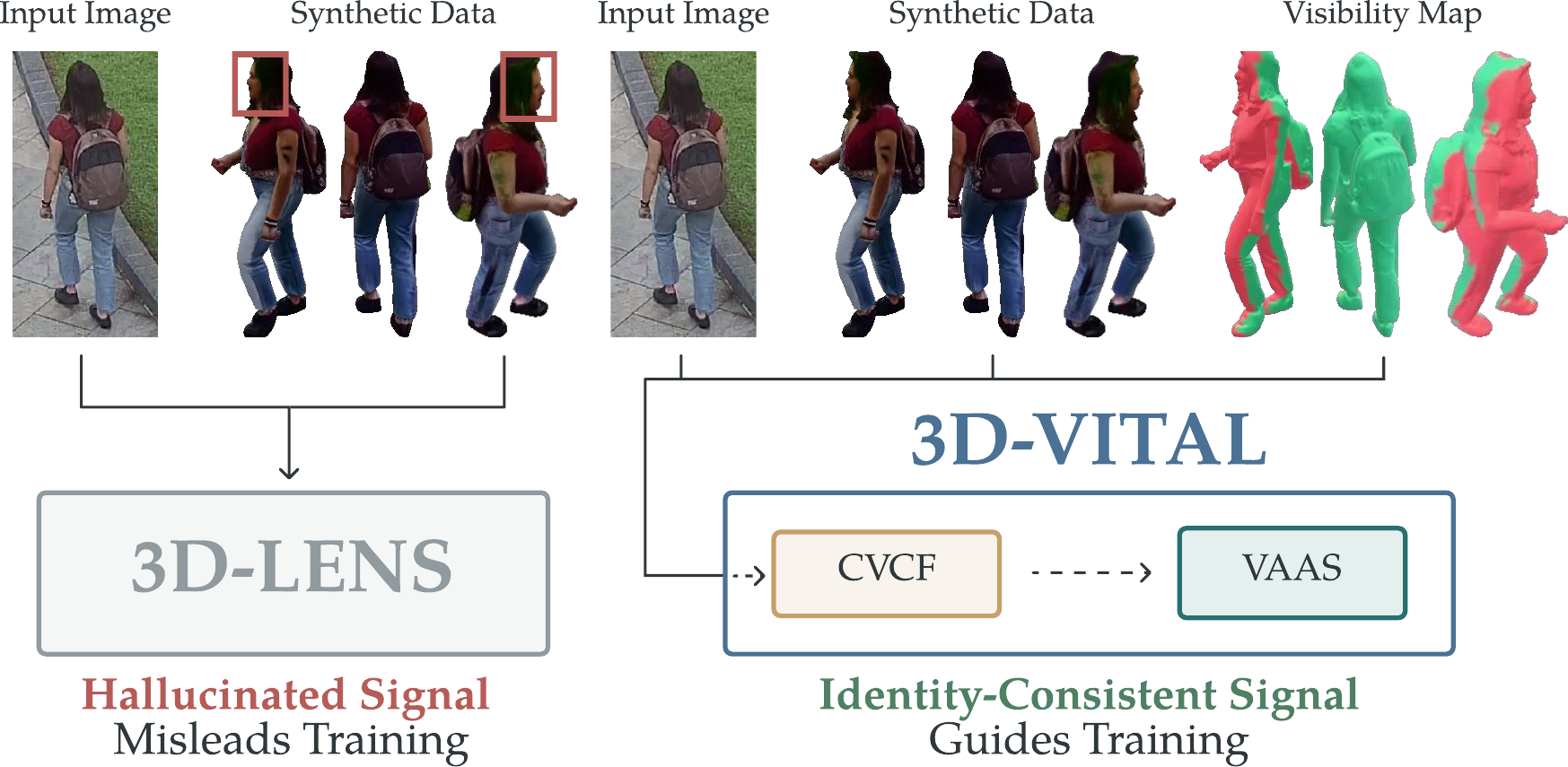}
        \caption{
            \textbf{Motivation of 3D-VITAL.}
            \textbf{Left:} Prior synthetic-real ReID pipelines such as 3D-LENS treat each synthetic sample as wholly reliable or wholly unreliable, which lets misleading identity signal from hallucinated regions (red rectangle) degrade training.
            \textbf{Right:} 3D-VITAL instead derives a per-pixel visibility map (visible regions as green, hallucinated as red) from the reconstructed mesh, separating observed regions from hallucinated ones.
            Cross-View Consistency Filtering (CVCF) uses this map to reject geometrically incorrect samples.
            Visibility-Aware Attention Supervision (VAAS) uses it to guide the model's attention toward observed regions during training.
        }
        \label{fig:gabstract}
    \end{figure}

    3D-LENS~\cite{grolleau20263dlens3dliftingbasedelevated} addresses this task by lifting single-view images into textured meshes using recent single-view 3D mesh reconstruction~\cite{zhao2025hunyuan3d}.
    It then renders novel views to simulate the missing viewpoint, providing cross-view supervision from single-view real data.
    However, this synthetic data is not always reliable for training for two reasons.
    When reconstruction fails to recover the correct geometry, the rendered novel views carry no usable identity signal.
    Even when the geometry is correct, some regions of the reconstructed object are not seen from the source view and are hallucinated by the reconstruction model.
    These models are trained on generic objects to produce visually plausible images, so hallucinated regions contain texture consistent with the object but without identity cues.
    Using such regions naively can degrade training, since the rendered novel views are only partially reliable.

    Working with unreliable training data is a well-studied problem in computer vision.
    Prior work either infers reliability from optimization dynamics~\cite{arpit2017closer, liu2020early} or derives it from external models~\cite{Miao_2019_ICCV, Gao_2020_CVPR, Wang_2020_CVPR}.
    However, neither approach applies in our setting, since 3D reconstruction models are explicitly optimized to produce visually plausible textures even in hallucinated regions, the rendered image alone provides no signal from which reliability could be estimated.
    Also, the 2D generative paradigms used in ReID, such as GANs and diffusion models, do not expose an explicit geometric signal from which a per-pixel reliability map could be derived.

    Closer to our setting, existing synthetic data pipelines for ReID do not exploit such a per-pixel signal.
    They either trust the synthetic data~\cite{Zheng_2017_ICCV, Zheng_2019_CVPR} or filter it at the image level~\cite{xiang2023less, niu2025synthesizing}.
    Both treat reliability as a property of the whole sample, so a novel view with correct geometry but partially hallucinated appearance is discarded entirely, wasting the valid identity signal it contains in its visible regions.

    In this paper, we address these limitations by computing a visibility map directly from the reconstructed mesh (\Cref{fig:gabstract}).
    This per-pixel labeling separates visible regions from hallucinated regions.
    Visible regions have direct correspondence to the source image and are therefore more likely to carry its identity signal, whereas hallucinated regions consist of pixels synthesized without source correspondence, potentially injecting identity-inconsistent noise.
    Building on this signal, we propose 3D-VITAL, which exploits visibility through a two-stage pipeline.
    The first stage, \textbf{Cross-View Consistency Filtering (CVCF)}, rejects source images whose meshes are geometrically incorrect.
    The second stage, \textbf{Visibility-Aware Attention Supervision (VAAS)}, operates on the surviving samples.
    We validate 3D-VITAL on three SV AG-ReID benchmarks (AG-ReID, AG-ReID.v2, MOO).

    In a nutshell, the main contributions of this paper are as follows:

    \begin{itemize}
        \item We identify that 3D reconstruction methods enable the computation of a visibility map, providing a dense reliability signal that prior generative methods do not naturally expose.
        \item We introduce Cross-View Consistency Filtering (CVCF) and a Visibility-Aware Attention Supervision (VAAS).
        CVCF scores reconstruction quality only over visible patches, while VAAS uses the visibility map to guide the model's attention toward reliable regions.
        \item We achieve state-of-the-art results on three SV AG-ReID benchmarks (AG-ReID, AG-ReID.v2, MOO), with ablations isolating the contribution of each component.
    \end{itemize}

    \section{Related Work}
    \label{sec:related_works}

    \paragraph{Estimating reliability.}
    When part of the training signal is unreliable, prior work that aims to increase robustness to input noise can be grouped into two broad categories.
    A first category infers reliability from the optimization process itself, exploiting the observation that networks fit clean samples before corrupted ones~\cite{arpit2017closer, liu2020early}.
    This property is exploited through peer-network disagreement~\cite{zhao2020unsupervised, yang2020asymmetric, yang2024robust, liu2024two}, clustering or Gaussian mixture models~\cite{SUN2024105244, zhong2024iclr, zhang2019self, qin2024noisy, yang2024robust}, or weighted losses based on uncertainty~\cite{SUN2024105244, Zheng_2019_CVPR, zhong2024iclr, jiang2021exploring}.
    A second category instead derives reliability from an external model or self-supervised proxy task, providing supervision in some regions of the image.
    The closest example to our setting is occluded person ReID, where external pose estimators produce heatmaps used as soft attention guidance~\cite{Miao_2019_ICCV, Gao_2020_CVPR, Wang_2020_CVPR, Somers_2023_WACV}, or where attention weights are learned through self-supervised augmentations~\cite{Sun_2019_CVPR, Wang_2022_CVPR, dutta2026latent, chen2021occlude}.
    However, in our setting, no such signal is available without explicit supervision since 3D reconstruction models produce plausible textures even in hallucinated regions, leaving no visual cue to distinguish reliable from unreliable patches, and no external model provides this partition for synthetic novel views.
    In contrast, the underlying 3D mesh exposes which regions are projected from the source view and which were hallucinated by the reconstruction model.
    Treating these regions differently during training lets us exploit synthetic data without being misled by the noise that hallucination introduces.

    \paragraph{Synthetic data for object re-identification.}
    Prior work using synthetic data in ReID treats reliability as a property of the whole sample, either by trusting all generated images or by scoring them at the image level.
    Early methods integrate GAN-synthesized images~\cite{Zheng_2017_ICCV,Zhong_2018_CVPR,Qian_2018_ECCV, ge2018fd, Zheng_2019_CVPR, chen2021joint}, diffusion-based generation~\cite{kim2024posediveposediversifiedaugmentationdiffusion, zhang2025viperson, li2025unleashing}, and more recently single-view 3D mesh reconstruction novel-view rendering~\cite{grolleau20263dlens3dliftingbasedelevated} into ReID training pipelines.
    These methods use synthetic data without specific reliability mechanisms, typically only smoothing labels to limit their influence relative to real datasets.
    However, many argue that curated subsets of synthetic data outperform larger uncurated data~\cite{xiang2023less, jiang2021exploring, yao2020simulating, he2020multi}, with recent ReID methods filtering synthetic data at the image level using ReID-based criteria~\cite{niu2025synthesizing}.
    Yet, in all of the above, reliability is a property of the full sample, with no mechanism to treat regions of an image independently.
    Reliability must instead be considered at the pixel level, separating visible regions that carry valid identity signal from hallucinated ones.
    To our knowledge, no prior synthetic-data pipeline for ReID exposes this distinction, and we are the first to exploit mesh-derived visibility as a per-pixel reliability signal for synthetic training data.

    \section{Methodology}

    3D-VITAL (\Cref{fig:gabstract}) is built upon 3D-LENS and extends it with three components that improve the use of synthetic data from single-view 3D mesh reconstruction.
    First, we compute a per-pixel visibility map that separates visible regions from hallucinated regions (\Cref{sec:visibility}).
    Second, Cross-View Consistency Filtering (CVCF) uses this visibility map to reject geometrically incorrect samples (\Cref{sec:filter}).
    Third, Visibility-Aware Attention Supervision (VAAS) operates on the surviving samples and uses the visibility map both as a per-patch attention target and as a per-sample weight (\Cref{sec:training}).
    These steps allow 3D-VITAL to exploit valid synthetic signals while mitigating the impact of unreliable regions.

    \subsection{Preliminaries}
    \label{sec:prelim}

    Following the baseline~\cite{grolleau20263dlens3dliftingbasedelevated}, each source image in the training set is lifted to a textured mesh and rendered from perturbed camera poses to provide cross-view supervision.
    Given a source image $I_i \in \mathbb{R}^{H \times W \times 3}$, a segmenter produces a foreground mask $M_i$, and a 3D reconstruction model $\mathcal{F}_{3D}$ produces a textured mesh $\mathcal{M}_i$.
    We calibrate a virtual camera $\Pi_i = [R_i \mid t_i]$ by maximizing the Intersection over Union (IoU) between the foreground mask $M_i$ and the silhouette of $\mathcal{M}_i$ under $\Pi_i$.
    This calibration ensures following viewpoint perturbations are relative to the source viewpoint, and that the visibility map (\Cref{sec:visibility}) is well-defined with respect to the source image.
    Then, $N$ novel views $I_i^{\text{syn}}$ are rendered, composited onto an inpainted background, and aligned in style via style transfer~\cite{chung2024styleid}, yielding the final synthetic views $I_i^{\text{sty}}$.

    The baseline uses a ViT backbone with a CLS token $t_{\text{id}}$ supervised by an identity classification loss $\mathcal{L}_{\text{id}}$, and a domain token $t_{\text{dom}}$ supervised by a binary cross-entropy loss $\mathcal{L}_{\text{dom}}$ that distinguishes real from synthetic data.
    We keep both losses as is.

    Our additions are the computation of a visibility map that is used with a Cross-View Consistency Filtering (CVCF), which removes incorrect samples before training, and a Visibility-Aware Attention Supervision (VAAS), which introduces a new attention loss term ($\mathcal{L}_{\text{vis}}$) and a weighted variant of the existing triplet term ($\mathcal{L}_{\text{tri}}^w$).

    \subsection{Visibility Map}
    \label{sec:visibility}

    We aim to compute a per-pixel map separating visible regions in the source image from regions hallucinated by the 3D reconstruction model.

    First, we build a visibility mesh $\mathcal{V}_i$ with the same topology as $\mathcal{M}_i$ but with a binary label per triangle.
    We employ ray casting from the source camera $\Pi_i$, marking the first triangle hit by each pixel ray as visible ($1$) and all other triangles as non-visible ($0$).
    Visible triangles are treated as more reliable because their appearance is constrained by the source image, although visibility does not guarantee a faithful reconstruction, while non-visible triangles carry texture hallucinated by $\mathcal{F}_{3D}$, since no observation constrained it.

    Second, for each rendered view $I_i^{\text{syn}} = \mathcal{R}(\mathcal{M}_i, \tilde{\Pi}_i)$, we render $\mathcal{V}_i$ under the same camera pose, yielding a binary visibility map $V_i \in \{0,1\}^{H \times W}$.
    Its pixels stay aligned with the foreground of $I_i^{\text{sty}}$, since background compositing and style transfer do not move the foreground.
    Spatial augmentations applied to $I_i^{\text{sty}}$ during training (horizontal flip, random crop) are applied identically to $V_i$ to preserve pixel correspondence.

    Third, we downsample $V_i$ to the $h \times w$ patch grid by average pooling over non-overlapping patches, yielding $\tilde{V}_i \in [0,1]^{h \times w}$.
    Each entry of $\tilde{V}_i$ is the fraction of visible pixels within its corresponding patch.
    This patch-resolution map is not fed to the network as input but used downstream as a per-sample reliability score for filtering (\Cref{sec:filter}) and a supervisory signal for attention (\Cref{sec:training}).

    \subsection{Cross-View Consistency Filtering}
    \label{sec:filter}

    \begin{figure}[tbp]
        \centering
        \begin{subfigure}[c]{0.74\linewidth}
            \includegraphics[width=\linewidth]{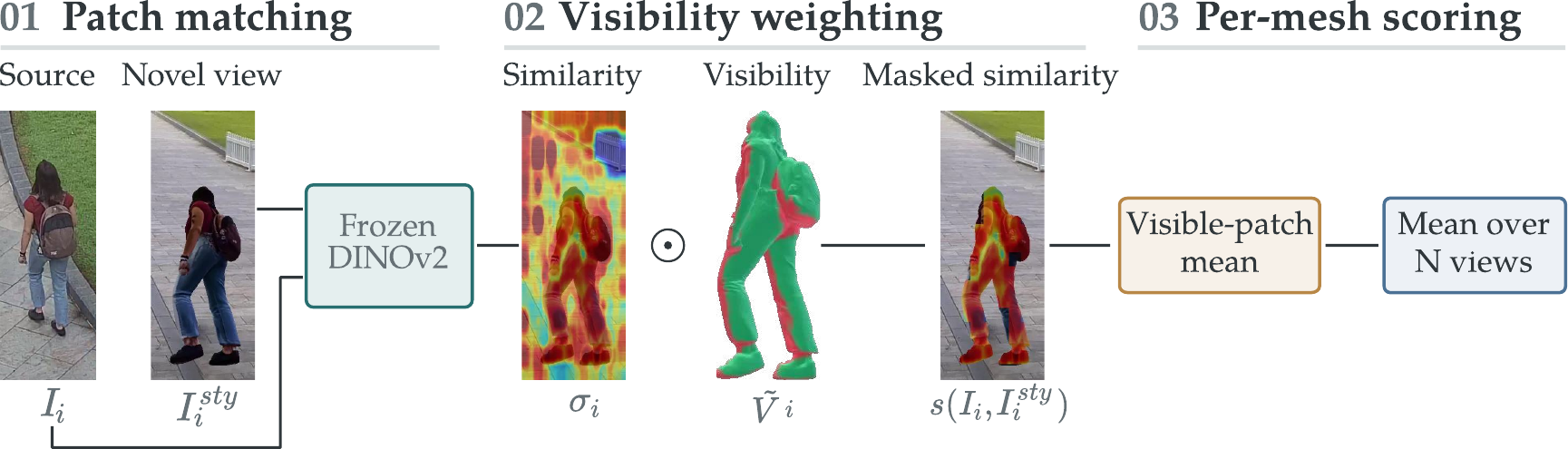}
            \caption{Cross-View Consistency Filtering (CVCF)}
            \label{fig:cvcf_pipeline}
        \end{subfigure}
        \hfill
        \begin{subfigure}[c]{0.23\linewidth}
            \includegraphics[width=\linewidth]{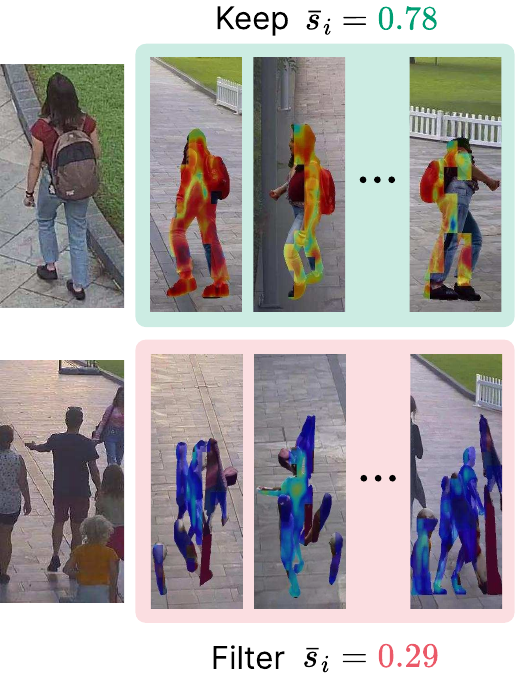}
            \caption{Kept and Filtered}
            \label{fig:cvcf_examples}
        \end{subfigure}
        \caption{
            \textbf{Cross-View Consistency Filtering (CVCF).}
            \textbf{(a)} Frozen DINOv2 features match each patch of a style-aligned novel view $I_i^{\text{sty}}$ to its most similar source-image patch, giving cosine similarities $\sigma_i$ (\Cref{eq:similarity}).
            The patch map $\tilde{V}_i$ masks these similarities to source-visible regions (\Cref{eq:score}), so hallucinated patches do not contribute.
            The visible-patch mean, averaged over the $N$ novel views, gives one consistency score $\bar{s}_i$ per source reconstruction (\Cref{eq:cvcf_aggregate}), used to keep the top samples.
            \textbf{(b)} A correct mesh (top) stays consistent across views in the visible regions, even when hallucinated regions are large, while an incorrect mesh (bottom) is inconsistent.
        }
        \label{fig:CVCF}
    \end{figure}

    Single-view 3D reconstruction may fail geometrically, producing novel views without identity signal.

    To filter these, we compute a consistency score $s$ based on the cosine similarity between DINOv2 features of the source image and its novel views, restricted to visible patches via $\tilde{V}_i$.
    DINOv2 patch features provide a robust signal for cross-view appearance consistency, while the visibility restriction ensures that only non-hallucinated regions contribute to the score.
    Unlike conventional image-level filters that completely discard heavily hallucinated views, this approach retains them.
    As long as the visible regions remain consistent with the source image, these reliable areas still provide a valuable cross-view signal that can be leveraged during training.

    Let $\phi(\cdot)$ denote patch features extracted by a frozen DINOv2 encoder.
    For each patch $j$ of a style-aligned novel view $I_i^{\text{sty}}$, we compute its maximum cosine similarity to any patch $k$ in the source image $I_i$,

    \begin{equation}
        \sigma_i^{(j)} = \max_{k} \cos(\phi(I_i^{\text{sty}})^{(j)}, \phi(I_i)^{(k)}).
        \label{eq:similarity}
    \end{equation}

    The image consistency score weights each patch similarity by its visibility:

    \begin{equation}
        s(I_i, I_i^{\text{sty}}) = \frac{\langle \tilde{V}_i, \sigma_i \rangle}{\|\tilde{V}_i\|_1}.
        \label{eq:score}
    \end{equation}

    The score is one-directional from novel view to source, since a bidirectional match would fail because novel viewpoints naturally occlude some source patches.
    We aggregate the $N$ novel views per source image by averaging their scores:

    \begin{equation}
        \bar{s}_i = \frac{1}{N}\sum_{n=1}^{N} s(I_i, I_i^{\text{sty},n}).
        \label{eq:cvcf_aggregate}
    \end{equation}

    This appearance-consistency score serves as a proxy for geometric correctness, since it does not enforce one-to-one or geometric correspondence.
    To filter out geometrically incorrect meshes, we keep the synthetic views of the $50\%$ of source images with the highest $\bar{s}_i$, while every real image remains in training.
    Sensitivity to this choice is reported in \Cref{tab:ablation_filter_tau}.

    \subsection{Visibility-Aware Attention Supervision}
    \label{sec:training}

    \begin{figure}[tbp]
        \centering
        \includegraphics[width=\linewidth]{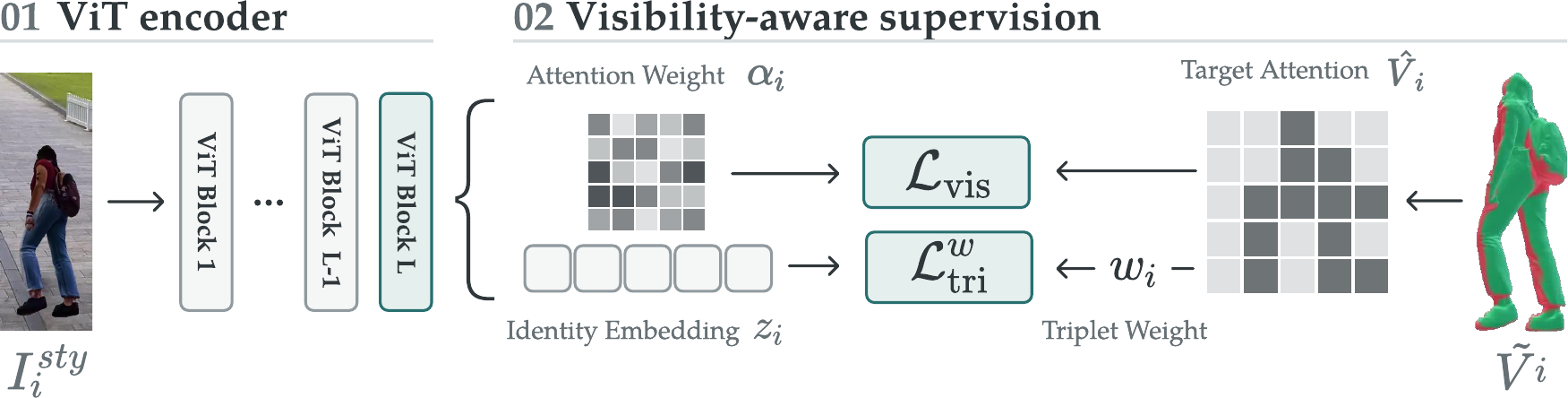}
        \caption{
            \textbf{Visibility-Aware Attention Supervision (VAAS).}
            A style-aligned novel view $I_i^{\text{sty}}$ is processed by a ViT.
            At the final transformer layer, we extract the attention weights $\alpha_i$ of the identity token $t_{\text{id}}$ over the image patches, averaged across heads.
            The visibility map $\tilde{V}_i$ provides two supervision signals.
            \textit{Top:} $\mathcal{L}_{\text{vis}}$ is the cross-entropy between $\alpha_i$ and the normalized visibility target $\hat{V}_i$ (\Cref{eq:vis}), pulling identity-token attention toward visible patches.
            \textit{Bottom:} the per-sample weight $w_i = \tfrac{1}{P}\sum_j \tilde{V}_i^{(j)}$ scales the anchor's contribution to $\mathcal{L}_{\text{tri}}^{w}$ (\Cref{eq:wtri}), reducing the influence of heavily hallucinated samples.
        }
        \label{fig:VAAS}
    \end{figure}

    Surviving samples contain the source image's identity signal in visible regions and misleading identity appearance in hallucinated ones.
    The visibility map exposes this partition directly, and we exploit it through two mechanisms.
    A visibility-aware identity loss ($\mathcal{L}_{\text{vis}}$) supervises the model to focus on visible regions, while a visibility-weighted triplet loss ($\mathcal{L}_{\text{tri}}^w$) reduces the impact of heavily hallucinated views on the overall optimization.
    This attention supervision acts as a regularization signal rather than a strict isolation, since final-layer patch features already mix information from other patches.

    \subsubsection*{Visibility-Aware Identity Loss}

    We supervise the attention weights of the CLS identity token at the final transformer layer to concentrate on visible patches, using the visibility map as a dense target distribution.
    We apply this supervision solely at the last transformer layer, allowing earlier layers to develop global context freely, and only to synthetic views.
    Hallucinated regions are kept in the input, since they still provide object context such as silhouette and pose (\Cref{tab:ablation_masked}).
    Ablation across the number of supervised layers is reported in \Cref{tab:ablation_n_layers}.

    In the final transformer layer, the identity token $t_{\text{id}}$ attends to itself, the domain token, and the $P$ image patches.
    We extract only the attention weights directed at the $P$ image patches, average them across all attention heads, and re-normalize them to sum to one, yielding the attention distribution $\alpha_i \in \mathbb{R}^P$.
    We also normalize the visibility map to a target probability distribution: $\hat{V}_i = \tilde{V}_i / (\epsilon + \sum_{j=1}^P \tilde{V}_i^{(j)})$, where $\epsilon$ is a small constant for numerical stability.
    We define the attention loss as the cross-entropy of this extracted attention against the visibility target distribution:

    \begin{equation}
        \mathcal{L}_{\text{vis}} = -\sum_{j=1}^{P} \hat{V}_i^{(j)} \log \alpha_i^{(j)}.
        \label{eq:vis}
    \end{equation}

    A comparison against alternative attention targets that ignore visible-vs-hallucinated information is reported in \Cref{tab:ablation_visibility_mode}.

    \subsubsection*{Visibility-Weighted Triplet Loss}

    The visibility map yields a per-sample reliability scalar $w_i = \frac{1}{P} \sum^P_{j=1} \tilde{V}_i^{(j)}$, representing the fraction of visible regions in the rendered view.
    We use $w_a$, obtained by setting $i=a$, to weight the sample's contribution as a triplet anchor.
    Given an anchor $a$, positive $p$, and negative $n$ drawn from a mini-batch, with embedding distance $D(\cdot,\cdot)$, the standard soft-margin triplet $\log\left(1+\exp\left(D(a, p) - D(a, n)\right)\right)$ is replaced by

    \begin{equation}
        \mathcal{L}_{\text{tri}}^{w} = w_a \cdot \log\left(1+\exp\left(D(a, p) - D(a, n)\right)\right).
        \label{eq:wtri}
    \end{equation}

    The weighted terms are normalized by the sum of the anchor weights in the batch, so anchors with low $w_a$ contribute proportionally less to the triplet loss.
    For real source images, we set $w_i \equiv 1$ and recover the standard term.

    \subsubsection*{Full Objective}
    \label{sec:obj}

    The total loss is:

    \begin{equation}
        \mathcal{L} = (\lambda_{\text{id}}\mathcal{L}_{\text{id}} + \lambda_{\text{dom}}\mathcal{L}_{\text{dom}}) + \lambda_{\text{tri}} \mathcal{L}_{\text{tri}}^{w}  + \lambda_{\text{vis}} \mathcal{L}_{\text{vis}},
        \label{eq:total}
    \end{equation}

    with $\lambda_{\text{id}}$ and $\lambda_{\text{dom}}$ set to $1$ and $0.5$ respectively as in the baseline, and with $\lambda_{\text{tri}}$, and $\lambda_{\text{vis}}$ being hyperparameters controlling the contribution of each term.
    Synthetic samples contribute to $\mathcal{L}_{\text{id}}$ and $\mathcal{L}_{\text{dom}}$ as in the baseline regardless of their visibility.
    Their influence is reduced only through the anchor weight of $\mathcal{L}_{\text{tri}}^{w}$, and $\mathcal{L}_{\text{vis}}$ is applied to synthetic views only.
    This joint optimization framework encourages the learning of discriminative representations that are resilient to synthetic hallucinations.

    \section{Experiments}

    \subsection{Implementation}

    \paragraph{Datasets and Evaluation protocol}.
    Throughout the paper, the notation $X \rightarrow Y$ denotes a query from view $X$ matched against a gallery of view $Y$, independent of which view the model was trained on.
    We evaluate 3D-VITAL on two pedestrian AG-ReID benchmarks, AG-ReID~\cite{nguyen2023aerial} and AG-ReID.v2~\cite{nguyen2024ag}, and on MOO~\cite{grolleau2026moomultivieworientedobservations}, a synthetic cattle dataset that tests generalization beyond pedestrians.
    AG-ReID contains 21,983 images and 388 identities from one ground camera and one UAV.
    AG-ReID.v2 contains 100,502 images and 1,615 identities across UAV, wearable, and CCTV platforms.
    MOO contains 128,000 images and 1,000 identities with elevations from $-25^\circ$ to $90^\circ$.
    We follow the Single-View AG-ReID protocol of~\cite{grolleau20263dlens3dliftingbasedelevated}, training each model on a single viewpoint and querying cross-viewpoint at inference.
    For AG-ReID and MOO, we train one model per view, yielding two training viewpoints per dataset (Aerial and Ground for AG-ReID, Top and Side for MOO).
    We evaluate both cross-view directions for each training view, giving four evaluations per dataset.
    For AG-ReID.v2, we train three models, one per view (Aerial, Wearable, and CCTV).
    The Wearable view acts as a second ground-like viewpoint and is evaluated against Aerial.
    We evaluate Aerial $\leftrightarrow$ CCTV and Aerial $\leftrightarrow$ Wearable in both directions for each training view, giving eight inference scenarios for AG-ReID.v2.

    \paragraph{Implementation Details}.
    We build on 3D-LENS~\cite{grolleau20263dlens3dliftingbasedelevated} with a ViT backbone and add our contributions incrementally to isolate their impact.
    We render synthetic views with a wider azimuth range than the baseline, since 3D-VITAL exploits viewpoint diversity without being harmed by the larger hallucinated regions it introduces (\Cref{subsubsec:view}).
    Images are resized to $256\times128$ and augmented with random cropping, horizontal flipping, and random erasing.
    The loss weights $\lambda_{\text{tri}}$ and $\lambda_{\text{vis}}$ from \Cref{eq:total} are both set to $1$.
    Sensitivity to these weights is reported in \Cref{tab:ablation_lambda_a,tab:ablation_lambda_d}.
    Training runs for 120 epochs with SGD (momentum $0.9$, weight decay $10^{-4}$) under cosine scheduling, with a learning rate of $8\times10^{-4}$ on AG-ReID and AG-ReID.v2 and $8\times10^{-3}$ on MOO.

    \subsection{Comparison with State-of-the-Art}

    We compare 3D-VITAL against prior methods on AG-ReID (\Cref{tab:sota_agreid_moo}), AG-ReID.v2 (\Cref{tab:sota_agreid2}), and MOO (\Cref{tab:sota_agreid_moo}).
    The baselines span standard ReID architectures (BoT~\cite{luo2019bag}, ViT~\cite{dosovitskiy2021image}, AGW~\cite{ye2021deep}, TransReID~\cite{he2021transreid}), viewpoint-agnostic methods (RotTrans~\cite{chen2022rotation}, VDT~\cite{Zhang_2024_CVPR}), large-scale pretraining (PASS~\cite{zhu2022pass}), generative adaptation (DCAC~\cite{li2025unleashing}), and the 3D-lifting baseline 3D-VITAL builds on (3D-LENS~\cite{grolleau20263dlens3dliftingbasedelevated}).
    All baselines use a ViT-Base backbone pretrained on ImageNet-21k~\cite{ridnik2021imagenet}, except PASS, which uses its own pretraining.

    3D-VITAL achieves the best mAP and Rank-1 across all four AG-ReID configurations (\Cref{tab:sota_agreid_moo}), reaching $53.2$ mAP and $62.6$ R-1 on Aerial $G\rightarrow A$, which is $+3.7$ mAP and $+4.9$ R-1 pp over 3D-LENS and far ahead of the best non-3D-lifting baseline (DCAC at $35.2$ / $42.7$).
    The gain is even more pronounced on MOO Side training $T \rightarrow S$ (\Cref{tab:sota_agreid_moo}), where 3D-VITAL reaches $47.8$ mAP and $99.6$ R-1, a $+15.7$ mAP pp jump over 3D-LENS.

    The pattern extends to AG-ReID.v2 (\Cref{tab:sota_agreid2}), where 3D-VITAL improves over 3D-LENS under CCTV training (\textit{e.g.}, $A\rightarrow C$: $46.3$ / $54.5$ vs. $41.4$ / $50.8$ mAP / R-1).
    Gains are smaller under Wearable training, where the higher image resolution ($222\times96$ vs.\ $187\times84$ for Aerial and $157\times59$ for CCTV) makes most reconstructions usable and the fixed $50\%$ filtering ratio overly aggressive.
    Even so, 3D-VITAL still outperforms every prior method other than 3D-LENS in these columns.

    These consistent gains demonstrate the effectiveness of 3D-VITAL's approach to improving the reliability of synthetic data from single-view 3D mesh reconstruction, enabling it to learn better cross-view representations.

    \begin{table}[tbp]
        \caption{
            \textbf{State-of-the-Art Comparison on AG-ReID and MOO.}
            mAP and Rank-1 (\%) on cross-view scenarios across Aerial (A), Ground (G) for AG-ReID, and Top (T), and Side (S) for MOO.
            Best in \textbf{bold}, second-best \underline{underlined}.
        }
        \label{tab:sota_agreid_moo}
        \centering

        \scriptsize
        \setlength{\tabcolsep}{3pt}

        \begin{tabular}{l c cccccccc cccccccc}
            \toprule
            & & \multicolumn{8}{c}{\textbf{AG-ReID}} & \multicolumn{8}{c}{\textbf{MOO}} \\

            \cmidrule(lr){3-10} \cmidrule(lr){11-18}

            &
            & \multicolumn{4}{c}{\textbf{Aerial Training}}
            & \multicolumn{4}{c}{\textbf{Ground Training}}
            & \multicolumn{4}{c}{\textbf{Top Training}}
            & \multicolumn{4}{c}{\textbf{Side Training}} \\

            \cmidrule(lr){3-6} \cmidrule(lr){7-10} \cmidrule(lr){11-14} \cmidrule(lr){15-18}

            \textbf{Methods} & \textbf{Year}
            & \multicolumn{2}{c}{$A \rightarrow G$}
            & \multicolumn{2}{c}{$G \rightarrow A$}
            & \multicolumn{2}{c}{$A \rightarrow G$}
            & \multicolumn{2}{c}{$G \rightarrow A$}
            & \multicolumn{2}{c}{$T \rightarrow S$}
            & \multicolumn{2}{c}{$S \rightarrow T$}
            & \multicolumn{2}{c}{$T \rightarrow S$}
            & \multicolumn{2}{c}{$S \rightarrow T$} \\

            \cmidrule(lr){3-4} \cmidrule(lr){5-6} \cmidrule(lr){7-8} \cmidrule(lr){9-10}
            \cmidrule(lr){11-12} \cmidrule(lr){13-14} \cmidrule(lr){15-16} \cmidrule(lr){17-18}

            & & mAP & R-1 & mAP & R-1 & mAP & R-1 & mAP & R-1
            & mAP & R-1 & mAP & R-1 & mAP & R-1 & mAP & R-1 \\
            \midrule

            BoT~\cite{luo2019bag} & 2019
            & 41.7 & 55.7 & 29.6 & 34.3 & 45.3 & 51.8 & 49.6 & 60.2
            & 37.1 & 95.9 & 39.1 & 43.5 & 19.3 & 43.1 & 19.4 & 94.2 \\

            ViT~\cite{dosovitskiy2021image} & 2020
            & 42.6 & 58.1 & 31.9 & 35.8 & 46.9 & 54.6 & 49.8 & 60.1
            & 36.5 & 93.9 & 37.8 & 41.6 & 19.8 & 43.5 & 20.0 & 94.8 \\

            AGW~\cite{ye2021deep} & 2021
            & 41.4 & 55.3 & 32.2 & 36.7 & 42.3 & 51.0 & 47.3 & 59.5
            & 36.1 & 89.2 & 37.8 & 44.0 & 13.5 & 40.8 & 13.8 & 76.6 \\

            TransReID~\cite{he2021transreid} & 2021
            & 32.7 & 46.4 & 23.5 & 25.2 & 44.2 & 50.7 & 49.7 & 62.6
            & 37.6 & 88.2 & 39.0 & 40.5 & 22.8 & 43.5 & 21.5 & 94.7 \\

            RotTrans~\cite{chen2022rotation} & 2022
            & 35.2 & 50.2 & 23.2 & 35.8 & 44.8 & 50.3 & 48.8 & 59.3
            & 41.4 & 99.2 & 41.6 & 42.4 & 30.7 & 43.2 & 30.7 & 97.4 \\

            PASS~\cite{zhu2022pass} & 2022
            & 39.1 & 52.2 & 26.0 & 28.6 & 47.7 & 56.2 & 47.8 & 58.0
            & 36.8 & 99.3 & 37.7 & 40.1 & 26.7 & 45.0 & 23.1 & 93.0 \\

            SBS~\cite{he2023fastreid} & 2023
            & 16.8 & 27.9 & 15.5 & 15.2 & 27.1 & 32.3 & 27.5 & 37.5
            & 14.1 & 63.3 & 18.9 & 19.8 & 0.8 & 1.1 & 0.9 & 0.9 \\

            DCT~\cite{li2023dc} & 2023
            & 22.0 & 35.8 & 14.0 & 14.2 & 36.2 & 43.8 & 40.6 & 52.3
            & 35.6 & 88.7 & 36.0 & 39.1 & 18.5 & 40.0 & 16.0 & 90.2 \\

            VDT~\cite{Zhang_2024_CVPR} & 2024
            & 39.0 & 53.3 & 31.7 & 36.1 & 48.3 & 56.0 & 51.4 & 63.2
            & 37.9 & 97.9 & 39.1 & 43.7 & 18.1 & 43.2 & 17.8 & 92.9 \\

            DCAC~\cite{li2025unleashing} & 2025
            & 43.4 & 60.0 & 35.2 & 42.7 & 49.7 & 61.6 & 52.9 & 66.4
            & 4.0 & 28.6 & 3.8 & 38.9 & 14.3 & 33.4 & 13.0 & 75.3 \\

            \midrule

            3D-LENS~\cite{grolleau20263dlens3dliftingbasedelevated} & 2026
            & \underline{49.8} & \underline{62.9} & \underline{49.5} & \underline{57.7} & \underline{53.3} & \underline{65.2} & \underline{55.1} & \underline{67.4}
            & \underline{51.4} & \textbf{100.0} & \underline{52.2} & \underline{55.6} & \underline{32.1} & \underline{57.9} & \underline{32.9} & \underline{99.0} \\

            \textbf{3D-VITAL} & \textbf{2026}
            & \textbf{52.1} & \textbf{66.1} & \textbf{53.2} & \textbf{62.6}
            & \textbf{56.0} & \textbf{67.0} & \textbf{57.9} & \textbf{68.6}
            & \textbf{57.0} & \textbf{100.0} & \textbf{53.2} & \textbf{60.0}
            & \textbf{47.8} & \textbf{99.6} & \textbf{45.1} & \textbf{99.6} \\

            \bottomrule
        \end{tabular}
    \end{table}

    \begin{table}[tbp]
        \caption{
            \textbf{State-of-the-Art Comparison on AG-ReID.v2.}
            mAP and Rank-1 (\%) on cross-view scenarios across Aerial (A), CCTV (C), and Wearable (W).
            Best in \textbf{bold}, second-best \underline{underlined}.
        }
        \label{tab:sota_agreid2}
        \centering

        \scriptsize
        \setlength{\tabcolsep}{3pt}

        \begin{tabular}{l c cccccccc cccc cccc}
            \toprule
            \multicolumn{18}{c}{\textbf{AG-ReID.v2}} \\

            \cmidrule(lr){1-18}

            &
            & \multicolumn{8}{c}{\textbf{Aerial Training}}
            & \multicolumn{4}{c}{\textbf{CCTV Training}}
            & \multicolumn{4}{c}{\textbf{Wearable Training}} \\

            \cmidrule(lr){3-10} \cmidrule(lr){11-14} \cmidrule(lr){15-18}

            \textbf{Methods} & \textbf{Year}
            & \multicolumn{2}{c}{$C \rightarrow A$}
            & \multicolumn{2}{c}{$A \rightarrow C$}
            & \multicolumn{2}{c}{$W \rightarrow A$}
            & \multicolumn{2}{c}{$A \rightarrow W$}
            & \multicolumn{2}{c}{$C \rightarrow A$}
            & \multicolumn{2}{c}{$A \rightarrow C$}
            & \multicolumn{2}{c}{$W \rightarrow A$}
            & \multicolumn{2}{c}{$A \rightarrow W$} \\

            \cmidrule(lr){3-4} \cmidrule(lr){5-6} \cmidrule(lr){7-8} \cmidrule(lr){9-10}
            \cmidrule(lr){11-12} \cmidrule(lr){13-14} \cmidrule(lr){15-16} \cmidrule(lr){17-18}

            & & mAP & R-1 & mAP & R-1 & mAP & R-1 & mAP & R-1
            & mAP & R-1 & mAP & R-1 & mAP & R-1 & mAP & R-1 \\
            \midrule

            BoT~\cite{luo2019bag} & 2019
            & 14.4 & 16.8 & 32.5 & 42.4 & 18.8 & 21.2 & 28.7 & 40.5
            & 42.3 & 52.6 & 34.5 & 40.0 & 31.6 & 38.0 & 25.1 & 28.9 \\

            ViT~\cite{dosovitskiy2021image} & 2020
            & 14.7 & 16.7 & 33.2 & 43.1 & 19.1 & 21.5 & 27.4 & 38.1
            & 41.8 & 51.8 & 35.0 & 41.0 & 31.0 & 37.1 & 23.9 & 27.9 \\

            AGW~\cite{ye2021deep} & 2021
            & 14.6 & 17.8 & 33.2 & 42.6 & 19.9 & 23.9 & 30.2 & 40.9
            & 38.3 & 47.4 & 31.5 & 38.2 & 32.4 & 39.1 & 25.3 & 31.2 \\

            TransReID~\cite{he2021transreid} & 2021
            & 11.5 & 12.4 & 25.3 & 34.0 & 15.8 & 18.4 & 23.8 & 34.1
            & 36.6 & 47.4 & 31.1 & 37.3 & 26.0 & 33.7 & 20.7 & 23.8 \\

            RotTrans~\cite{chen2022rotation} & 2022
            & 15.2 & 16.6 & 32.6 & 42.1 & 20.7 & 21.8 & 30.4 & 42.5
            & 41.6 & 52.4 & 34.8 & 39.7 & 31.6 & 40.4 & 28.7 & 31.2 \\

            PASS~\cite{zhu2022pass} & 2022
            & 16.6 & 21.1 & 32.4 & 43.6 & 16.2 & 19.4 & 26.3 & 38.8
            & 40.5 & 50.5 & 36.5 & 42.4 & 31.0 & 39.8 & 25.8 & 29.8 \\

            SBS~\cite{he2023fastreid} & 2023
            & 2.5 & 2.7 & 3.8 & 5.1 & 3.2 & 3.3 & 2.7 & 2.8
            & 14.3 & 21.5 & 12.6 & 14.3 & 3.8 & 5.0 & 3.5 & 3.0 \\

            DCT~\cite{li2023dc} & 2023
            & 5.8 & 6.2 & 12.6 & 20.2 & 10.4 & 12.1 & 12.8 & 22.1
            & 31.6 & 43.4 & 26.0 & 30.6 & 17.9 & 26.6 & 15.8 & 17.8 \\

            VDT~\cite{Zhang_2024_CVPR} & 2024
            & 14.6 & 17.3 & 32.5 & 42.5 & 18.3 & 21.4 & 26.4 & 37.6
            & 41.1 & 51.7 & 33.9 & 39.8 & 29.6 & 37.2 & 23.3 & 26.4 \\

            DCAC~\cite{li2025unleashing} & 2025
            & 14.4 & 18.1 & 34.9 & 47.9 & 19.8 & 23.5 & 32.6 & 46.8
            & 45.6 & 57.9 & 39.0 & 48.4 & 31.9 & 41.6 & 30.1 & 35.8 \\

            \midrule

            3D-LENS~\cite{grolleau20263dlens3dliftingbasedelevated} & 2026
            & \underline{37.8} & \underline{47.1} & \underline{42.8} & \underline{55.7} & \underline{28.6} & \underline{35.8} & \underline{34.3} & \underline{47.1}
            & \underline{47.2} & \underline{58.5} & \underline{41.4} & \underline{50.8} & \textbf{38.3} & \textbf{46.2} & \textbf{38.9} & \textbf{47.9} \\

            \textbf{3D-VITAL} & \textbf{2026}
            & \textbf{39.9} & \textbf{47.3} & \textbf{43.9} & \textbf{56.5} & \textbf{31.1} & \textbf{37.1} & \textbf{35.2} & \textbf{48.3}
            & \textbf{50.4} & \textbf{61.1} & \textbf{46.3} & \textbf{54.5} & \underline{37.9} & \underline{45.8} & \textbf{38.9} & \underline{46.9} \\

            \bottomrule
        \end{tabular}
    \end{table}

    \subsection{Ablation Study}

    \paragraph{Component Analysis}.
    To assess the individual contribution of each 3D-VITAL component, we conduct an ablation study on the AG-ReID dataset.
    Following the same synthetic data generation as the baseline~\cite{grolleau20263dlens3dliftingbasedelevated}, we report the contribution of each 3D-VITAL component in \Cref{tab:comp_components}.

    Each component contributes independently: on Aerial $A\rightarrow G$, $\text{VAAS}_{\text{vis}}$ alone yields $+3.2$ mAP pp (Index 1), $\text{VAAS}_{\text{tri}}$ $+0.9$ pp (Index 2), and CVCF the largest standalone gain at $+5.1$ pp (Index 4) since broken meshes are the most damaging samples.
    Combined (Index 5), the three components reach $+6.5$ mAP pp under Aerial training and $+6.3$ under Ground training.
    CVCF removes broken geometry upstream while VAAS handles partial visibility within the surviving samples and together enable the model to learn more effectively from the synthetic data.

    \begin{table}[tbp]
        \caption{
            \textbf{Component Ablation on AG-ReID.}
            mAP and Rank-1 (\%) on Aerial $\rightarrow$ Ground and Ground $\rightarrow$ Aerial, with gains over Index 0 in green.
            Best in \textbf{bold}.
        }
        \label{tab:comp_components}
        \centering

        \scriptsize
        \setlength{\tabcolsep}{1.5pt}

        \begin{tabular}{c ccc cccc cccc}
            \toprule
            \multicolumn{1}{c}{\textbf{Index}}
            & \multicolumn{3}{c}{\textbf{Components}}
            & \multicolumn{4}{c}{\textbf{AG-ReID Aerial}}
            & \multicolumn{4}{c}{\textbf{AG-ReID Ground}} \\

            \cmidrule(lr){2-4} \cmidrule(lr){5-8} \cmidrule(lr){9-12}

            & \multirow{2.5}{*}{$\text{VAAS}_{\text{vis}}$}
            & \multirow{2.5}{*}{$\text{VAAS}_{\text{tri}}$}
            & \multirow{2.5}{*}{CVCF}
            & \multicolumn{2}{c}{$A \rightarrow G$}
            & \multicolumn{2}{c}{$G \rightarrow A$}
            & \multicolumn{2}{c}{$A \rightarrow G$}
            & \multicolumn{2}{c}{$G \rightarrow A$} \\

            \cmidrule(lr){5-6} \cmidrule(lr){7-8} \cmidrule(lr){9-10} \cmidrule(lr){11-12}

            & & & & mAP & R-1 & mAP & R-1 & mAP & R-1 & mAP & R-1 \\
            \midrule

            0 & & &
            & 45.6 & 61.2 & 46.1 & 56.4
            & 49.7 & 61.6 & 51.3 & 62.9 \\

            1 & \checkmark & &
            & \resPos{48.8}{3.2} & \resPos{63.9}{2.6} & \resPos{50.0}{3.8} & \resPos{59.6}{3.1}
            & \resPos{52.2}{2.5} & \resPos{63.3}{1.7} & \resPos{54.6}{3.3} & \resPos{65.8}{2.9} \\

            2 & & \checkmark &
            & \resPos{46.5}{0.9} & \resPos{62.2}{0.9} & \resPos{46.9}{0.8} & \resPos{56.9}{0.4}
            & \resPos{50.6}{0.9} & \resPos{62.9}{1.3} & \resPos{52.9}{1.6} & \resPos{63.6}{0.7} \\

            3 & \checkmark & \checkmark &
            & \resPos{49.2}{3.6} & \resPos{64.2}{3.0} & \resPos{49.5}{3.3} & \resPos{58.6}{2.2}
            & \resPos{53.0}{3.3} & \resPos{64.5}{2.9} & \resPos{55.5}{4.3} & \resPos{67.6}{4.7} \\

            4 & & & \checkmark
            & \resPos{50.7}{5.1} & \resPos{65.6}{4.4} & \resPos{52.0}{5.9} & \resPos{60.2}{3.8}
            & \resPos{54.0}{4.3} & \resPos{65.1}{3.5} & \resPos{56.3}{5.0} & \resPos{67.4}{4.5} \\

            5 & \checkmark & \checkmark & \checkmark
            & \resPosBold{52.1}{6.5} & \resPosBold{66.1}{4.9} & \resPosBold{53.2}{7.1} & \resPosBold{62.6}{6.2}
            & \resPosBold{56.0}{6.3} & \resPosBold{67.0}{5.4} & \resPosBold{57.9}{6.7} & \resPosBold{68.6}{5.7} \\

            \bottomrule
        \end{tabular}
    \end{table}

    \paragraph{Novel View Augmentation Analysis}.
    \label{subsubsec:view}
    \Cref{tab:ablation_azimuth} reports the effect of the azimuth augmentation range on AG-ReID, with and without 3D-VITAL.
    Without 3D-VITAL, performance peaks at the moderate $\pm 30^\circ$ range and degrades at $\pm 60^\circ$, dropping by $1.3$ mAP pp on Aerial $A\rightarrow G$ relative to $\pm 30^\circ$, since wider perturbations expose the baseline to rendered views with larger hallucinated regions.
    Once 3D-VITAL components are enabled, the accuracy increases instead, reaching $52.1$ mAP on Aerial $A \rightarrow G$ at $\pm 60^\circ$, a gain of $+2.3$ mAP pp over the same setting at $\pm 30^\circ$.
    This shift is consistent with the role of CVCF and VAAS, since geometrically broken meshes are filtered out and partial visibility is handled during training, the model can exploit the larger viewpoint diversity offered by wider azimuth perturbations rather than being harmed by it.

    \begin{table}[tbp]
        \caption{
            \textbf{Azimuth Range Ablation on AG-ReID.}
            mAP and Rank-1 (\%) on Aerial $\rightarrow$ Ground and Ground $\rightarrow$ Aerial.
            Best in \textbf{bold}, second-best \underline{underlined}.
        }
        \label{tab:ablation_azimuth}
        \centering

        \scriptsize

        \begin{tabular}{l c cccc cccc}
            \toprule
            & & \multicolumn{4}{c}{\textbf{Aerial Training}} & \multicolumn{4}{c}{\textbf{Ground Training}} \\
            \cmidrule(lr){3-6} \cmidrule(lr){7-10}
            \textbf{Setting} & \textbf{Azimuth}
            & \multicolumn{2}{c}{$A \rightarrow G$}
            & \multicolumn{2}{c}{$G \rightarrow A$}
            & \multicolumn{2}{c}{$A \rightarrow G$}
            & \multicolumn{2}{c}{$G \rightarrow A$} \\
            \cmidrule(lr){3-4} \cmidrule(lr){5-6} \cmidrule(lr){7-8} \cmidrule(lr){9-10}
            &                & mAP           & R-1           & mAP           & R-1           & mAP           & R-1           & mAP           & R-1           \\
            \midrule
            \multirow{3}{*}{\textit{Baseline}}
            & $\pm 15^\circ$ & 46.5          & 61.7          & 47.2          & 57.2          & 50.0          & 62.3          & 51.1          & 63.3          \\
            & $\pm 30^\circ$ & \textbf{46.9} & \textbf{62.6} & \textbf{47.9} & \textbf{57.9} & \textbf{50.6} & \textbf{62.6} & \textbf{52.9} & \textbf{64.3} \\
            & $\pm 60^\circ$ & 45.6          & 61.2          & 46.1          & 56.4          & 49.7          & 61.6          & 51.3          & 62.9          \\
            \midrule
            \multirow{3}{*}{\textit{3D-VITAL}}
            & $\pm 15^\circ$ & 48.7          & 63.0          & 50.0          & 58.1          & 54.9          & 65.3          & 56.1          & 67.0          \\
            & $\pm 30^\circ$ & 49.8          & 64.8          & 51.0          & 58.3          & 55.2          & 66.4          & 56.5          & 66.9          \\
            & $\pm 60^\circ$ & \textbf{52.1} & \textbf{66.1} & \textbf{53.2} & \textbf{62.6} & \textbf{56.0} & \textbf{67.0} & \textbf{57.9} & \textbf{68.6} \\
            \bottomrule
        \end{tabular}
    \end{table}

    \paragraph{Filtering Strategy Analysis}.
    \Cref{tab:ablation_filtering} compares filtering strategies on AG-ReID with and without VAAS.
    CLIP-based filters underperform because image-to-image scoring does not capture cross-view identity consistency and image-to-text only verifies subject presence.
    DINOv2 variants improve on this, but both let hallucinated regions corrupt the score.
    Diff-Person~\cite{niu2025synthesizing}, with its ReID-CTF module, is the strongest non-visibility baseline but introduces a paradox.
    ReID models penalize the extreme-view samples that are precisely the reason for generating novel views in the first place.
    Restricting the score to visible patches (ours) avoids both issues and leads to the best performance across all settings.

    The interaction with VAAS further highlights the importance of visibility-aware filtering.
    Without VAAS, our filter leads by a modest margin ($52.0$ vs $51.2$ mAP for ReID-CTF on Aerial $G \rightarrow A$).
    With VAAS, the gap widens to between $1.2$ and $1.5$ mAP and is consistent across all eight settings.
    Image-level filters cannot exploit this interaction because they discard the partially-hallucinated samples that VAAS is designed to learn from.

    \begin{table}[tbp]
        \caption{
            \textbf{Filtering Strategy Ablation on AG-ReID.}
            mAP and Rank-1 (\%) on Aerial $\rightarrow$ Ground and Ground $\rightarrow$ Aerial.
            Best per column within each sub-block in \textbf{bold}, second-best \underline{underlined}.
        }
        \label{tab:ablation_filtering}
        \centering

        \scriptsize
        \setlength{\tabcolsep}{4pt}
        \renewcommand{\arraystretch}{1.05}

        \begin{tabular}{l l cc cc cc cc}
            \toprule
            & & \multicolumn{4}{c}{Aerial Training} & \multicolumn{4}{c}{Ground Training} \\
            \cmidrule(lr){3-6} \cmidrule(lr){7-10}
            \textbf{Setting} & \textbf{Filter}
            & \multicolumn{2}{c}{$A \rightarrow G$}
            & \multicolumn{2}{c}{$G \rightarrow A$}
            & \multicolumn{2}{c}{$A \rightarrow G$}
            & \multicolumn{2}{c}{$G \rightarrow A$} \\
            \cmidrule(lr){3-4} \cmidrule(lr){5-6} \cmidrule(lr){7-8} \cmidrule(lr){9-10}
            & & mAP & R-1 & mAP & R-1 & mAP & R-1 & mAP & R-1 \\
            \midrule
            \multirow{7}{*}{\textit{\shortstack[l]{Without\\VAAS}}}
            & No filter
            & 45.6 & 61.2 & 46.1 & 56.4 & 49.7 & 61.6 & 51.3 & 62.9 \\
            & CLIP I-I
            & 46.6 & 61.0 & 47.1 & 57.1 & 49.0 & 61.6 & 51.8 & 64.1 \\
            & CLIP I-T
            & 46.9 & 61.7 & 47.7 & 57.8 & 51.8 & 63.5 & 54.8 & 65.8 \\
            & DINOv2 CLS
            & 48.5 & 64.2 & 50.4 & \underline{59.5} & 52.0 & 63.3 & 54.3 & 65.8 \\
            & DINOv2 w/o Vis
            & 49.8 & 64.4 & \underline{51.2} & 59.2 & \underline{52.8} & 63.8 & \underline{55.3} & \underline{65.9} \\
            & ReID-CTF~\cite{niu2025synthesizing}
            & \underline{50.0} & \underline{64.8} & \underline{51.2} & 59.4 & 52.7 & \underline{64.9} & 55.1 & 65.8 \\
            & DINOv2 Vis (ours)
            & \textbf{50.7} & \textbf{65.6} & \textbf{52.0} & \textbf{60.2} & \textbf{54.0} & \textbf{65.1} & \textbf{56.3} & \textbf{67.4} \\
            \midrule
            \multirow{6}{*}{\textit{\shortstack[l]{With\\VAAS}}}
            & CLIP I-I
            & 49.8 & 64.1 & 50.8 & 59.3 & 52.4 & 63.4 & 55.6 & 65.3 \\
            & CLIP I-T
            & 49.6 & 63.9 & 50.7 & 59.1 & 52.9 & 63.0 & 55.6 & 65.3 \\
            & DINOv2 CLS
            & 50.5 & 65.0 & 51.5 & 60.3 & 54.2 & 65.3 & 56.6 & \underline{66.7} \\
            & DINOv2 w/o Vis
            & 50.3 & 64.7 & 51.3 & \underline{60.7} & \underline{54.8} & 65.4 & \underline{55.9} & 66.0 \\
            & ReID-CTF~\cite{niu2025synthesizing}
            & \underline{50.7} & \underline{64.9} & \underline{51.7} & 60.4 & \underline{54.8} & \underline{65.5} & \underline{55.9} & 66.5 \\
            & DINOv2 Vis (ours)
            & \textbf{52.1} & \textbf{66.1} & \textbf{53.2} & \textbf{62.6} & \textbf{56.0} & \textbf{67.0} & \textbf{57.9} & \textbf{68.6} \\
            \bottomrule
        \end{tabular}
    \end{table}

    \section{Conclusion and Limitations}
    \label{sec:conclusion}

    \paragraph{Conclusion.} We presented 3D-VITAL, which turns the visibility map exposed by single-view 3D mesh reconstruction into a training signal for cross-view ReID.
    Cross-View Consistency Filtering (CVCF) rejects samples with incorrect geometry but retains those with correct geometry and partial hallucination.
    Our Visibility-Aware Attention Supervision (VAAS) then supervises the CLS-attention and reweights triplet anchors, so the model learns from the visible regions of the surviving samples without being misled by hallucinated ones.
    Together, the two stages let 3D-VITAL train on a wider viewpoint range than the baseline and set a new state of the art on AG-ReID, AG-ReID.v2, and MOO.

    \paragraph{Limitations.}
    3D-VITAL inherits the failure modes of its upstream components.
    Low resolution, severe occlusion, or multiple individuals per image may degrade reconstruction quality, causing CVCF to discard samples and reduce the augmented training set.
    While effective for isolated noisy samples, this filtering creates a bottleneck when poor reconstructions are systematic.
    Additionally, the fixed filtering ratio introduces a related issue.
    On the high-resolution Wearable split of AG-ReID.v2, discarding the bottom $50\%$ of samples is overly aggressive because the average reconstruction quality is high.
    Finally, objects with atypical geometries or fine details that were not seen during the 3D reconstruction model's training distribution often yield degenerate meshes and further limit the applicability of 3D-VITAL to such cases.
    Finally, as with any ReID method, improved aerial-ground matching could be misused for surveillance of individuals without consent, and deployments should comply with applicable privacy regulations.

    \begin{ack}
        This publication was made possible by the use of the FactoryIA supercomputer, financially supported by the Ile-De-France Regional Council.
        The authors declare no competing interests.
    \end{ack}

    \medskip
    \small

    \bibliographystyle{unsrt}
    \bibliography{main}

@String(CVPR  = {IEEE Conf. Comput. Vis. Pattern Recog.})

@String(WACV  = {IEEE Winter Conf. Appl. Comput. Vis.})

@String(ICCV  = {Int. Conf. Comput. Vis.})

@String(ECCV  = {Eur. Conf. Comput. Vis.})

@String(NeurIPS = {Adv. Neural Inform. Process. Syst.})

@String(ICLR  = {Int. Conf. Learn. Represent.})

@String(CVPRW = {IEEE Conf. Comput. Vis. Pattern Recog. Worksh.})

@String(AAAI  = {AAAI})

@String(ICME  = {Int. Conf. Multimedia and Expo})

@String(TMM   = {IEEE Trans. Multimedia})

@String(TIFS = {IEEE Trans. Inf. Forensics Secur.})

@String(ACMMM = {ACM Int. Conf. Multimedia})

@String(RS = {Remote Sensing})

@String(Sensors = {Sensors})

@String( CVPR = {CVPR})

@String( WACV = {WACV})

@String( ICCV = {ICCV})

@String( ECCV = {ECCV})

@String( NeurIPS = {NeurIPS})

@String( ICLR = {ICLR})

@String( CVPRW = {CVPRW})

@String( AAAI = {AAAI})

@String( ICME =    {ICME})

@String( TMM =    {IEEE TMM})

@String( TIFS = {IEEE TIFS})

@String( ACMMM = {ACM MM})

@String( RS = {Remote Sens.})

@String( Sensors = {Sensors})

@inproceedings{nguyen2023aerial,
    title = {Aerial-ground person re-id},
    author = {Nguyen, Huy and Nguyen, Kien and Sridharan, Sridha and Fookes, Clinton},
    booktitle = ICME,
    pages = {2585--2590},
    year = {2023}
}

@article{nguyen2024ag,
    title = {AG-ReID. v2: Bridging aerial and ground views for person re-identification},
    author = {Nguyen, Huy and Nguyen, Kien and Sridharan, Sridha and Fookes, Clinton},
    journal = TIFS,
    volume = {19},
    pages = {2896--2908},
    year = {2024},
    publisher = {IEEE}
}

@inproceedings{Zhang_2024_CVPR,
  author    = {Zhang, Quan and Wang, Lei and Patel, Vishal M. and Xie, Xiaohua and Lai, Jianhuang},
  title     = {View-decoupled Transformer for Person Re-identification under Aerial-ground Camera Network},
  booktitle = {Proceedings of the IEEE/CVF Conference on Computer Vision and Pattern Recognition (CVPR)},
  pages     = {22000--22009},
  year      = {2024},
  doi       = {10.1109/CVPR52733.2024.02077}
}

@inproceedings{wang2025secap,
    title = {SeCap: Self-Calibrating and Adaptive Prompts for Cross-view Person Re-Identification in Aerial-Ground Networks},
    author = {Wang, Shining and Wang, Yunlong and Wu, Ruiqi and Jiao, Bingliang and Wang, Wenxuan and Wang, Peng},
    booktitle = CVPR,
    pages = {22119--22128},
    year = {2025}
}

@article{zhang2020person,
    title = {Person re-identification in aerial imagery},
    author = {Zhang, Shizhou and Zhang, Qi and Yang, Yifei and Wei, Xing and Wang, Peng and Jiao, Bingliang and Zhang, Yanning},
    journal = TMM,
    volume = {23},
    pages = {281--291},
    year = {2020}
}

@article{shang2025aerial,
    title = {Aerial-Ground Cross-View Vehicle Re-Identification: A Benchmark Dataset and Baseline},
    author = {Shang, Linzhi and Min, Chen and Wang, Juan and Xiao, Liang and Zhao, Dawei and Nie, Yiming},
    journal = RS,
    volume = {17},
    number = {15},
    pages = {2653},
    year = {2025},
    publisher = {MDPI}
}

@inproceedings{grolleau20263dlens3dliftingbasedelevated,
  author    = {Grolleau, William and Sabourin, Astrid and Lapouge, Guillaume and Achard, Catherine},
  title     = {3D-LENS: A 3D Lifting-based Elevated Novel-view Synthesis Method for Single-View Aerial-Ground Re-Identification},
  booktitle = {Proceedings of the European Conference on Computer Vision (ECCV)},
  pages     = {205--221},
  year      = {2026},
  doi       = {10.1007/978-3-032-36846-1_12}
}

@inproceedings{kim2024posediveposediversifiedaugmentationdiffusion,
  author    = {Inès Hyeonsu Kim and JoungBin Lee and Woojeong Jin and Soowon Son and Kyusun Cho and Junyoung Seo and Min-Seop Kwak and Seokju Cho and JeongYeol Baek and Byeongwon Lee and Seungryong Kim},
  title     = {Pose-dIVE: Pose-Diversified Augmentation for Person Re-Identification},
  booktitle = {Proceedings of the IEEE/CVF Conference on Computer Vision and Pattern Recognition (CVPR) Findings},
  pages     = {8640--8650},
  year      = {2026}
}

@inproceedings{zhang2025viperson,
    title = {VIPerson: Flexibly Generating Virtual Identity for Person Re-Identification},
    author = {Zhang, Xiao-Wen and Zhang, Delong and Peng, Yi-Xing and Ouyang, Zhi and Meng, Jingke and Zheng, Wei-Shi},
    booktitle = ICCV,
    pages = {23374--23384},
    year = {2025}
}

@article{niu2025synthesizing,
  author    = {Niu, Ke and Yu, Haiyang and Qian, Xuelin and Fu, Teng and Li, Bin and Xue, Xiangyang},
  title     = {Synthesizing efficient data with diffusion models for person re-identification pre-training},
  journal   = {Machine Learning},
  volume    = {114},
  number    = {3},
  pages     = {66},
  year      = {2025},
  doi       = {10.1007/s10994-024-06663-3}
}

@inproceedings{chen2021joint,
    title = {Joint generative and contrastive learning for unsupervised person re-identification},
    author = {Chen, Hao and Wang, Yaohui and Lagadec, Benoit and Dantcheva, Antitza and Bremond, Francois},
    booktitle = CVPR,
    pages = {2004--2013},
    year = {2021}
}

@article{li2025unleashing,
    title = {Unleashing the Potential of Pre-Trained Diffusion Models for Generalizable Person Re-Identification},
    author = {Li, Jiachen and Gong, Xiaojin},
    journal = Sensors,
    volume = {25},
    number = {2},
    pages = {552},
    year = {2025}
}

@InProceedings{Zhong_2018_CVPR,
    author = {Zhong, Zhun and Zheng, Liang and Zheng, Zhedong and Li, Shaozi and Yang, Yi},
    title = {Camera Style Adaptation for Person Re-Identification},
    booktitle = {Proceedings of the IEEE Conference on Computer Vision and Pattern Recognition (CVPR)},
    month = {June},
    year = {2018}
}

@InProceedings{Qian_2018_ECCV,
    author = {Qian, Xuelin and Fu, Yanwei and Xiang, Tao and Wang, Wenxuan and Qiu, Jie and Wu, Yang and Jiang, Yu-Gang and Xue, Xiangyang},
    title = {Pose-Normalized Image Generation for Person Re-identification},
    booktitle = {Proceedings of the European Conference on Computer Vision (ECCV)},
    month = {September},
    year = {2018}
}

@inproceedings{ge2018fd,
  author    = {Ge, Yixiao and Li, Zhuowan and Zhao, Haiyu and Yin, Guojun and Yi, Shuai and Wang, Xiaogang and Li, Hongsheng},
  title     = {FD-GAN: Pose-guided Feature Distilling GAN for Robust Person Re-identification},
  booktitle = {Advances in Neural Information Processing Systems (NeurIPS)},
  pages     = {1230--1241},
  year      = {2018}
}

@InProceedings{Zheng_2017_ICCV,
    author = {Zheng, Zhedong and Zheng, Liang and Yang, Yi},
    title = {Unlabeled Samples Generated by GAN Improve the Person Re-Identification Baseline in Vitro},
    booktitle = {Proceedings of the IEEE International Conference on Computer Vision (ICCV)},
    month = {Oct},
    year = {2017}
}

@InProceedings{Zheng_2019_CVPR,
    author = {Zheng, Zhedong and Yang, Xiaodong and Yu, Zhiding and Zheng, Liang and Yang, Yi and Kautz, Jan},
    title = {Joint Discriminative and Generative Learning for Person Re-Identification},
    booktitle = {Proceedings of the IEEE/CVF Conference on Computer Vision and Pattern Recognition (CVPR)},
    month = {June},
    year = {2019}
}

@article{xiang2023less,
  author    = {Xiang, Suncheng and Qian, Dahong and Guan, Mengyuan and Yan, Binjie and Liu, Ting and Fu, Yuzhuo and You, Guanjie},
  title     = {Less Is More: Learning from Synthetic Data with Fine-Grained Attributes for Person Re-Identification},
  journal   = {ACM Transactions on Multimedia Computing, Communications, and Applications},
  volume    = {19},
  number    = {5s},
  pages     = {173:1--173:20},
  year      = {2023},
  doi       = {10.1145/3588441}
}

@article{liu2020early,
    title = {Early-learning regularization prevents memorization of noisy labels},
    author = {Liu, Sheng and Niles-Weed, Jonathan and Razavian, Narges and Fernandez-Granda, Carlos},
    journal = {Advances in neural information processing systems},
    volume = {33},
    pages = {20331--20342},
    year = {2020}
}

@inproceedings{arpit2017closer,
    title = {A closer look at memorization in deep networks},
    author = {Arpit, Devansh and Jastrz{\k{e}}bski, Stanis{\l}aw and Ballas, Nicolas and Krueger, David and Bengio, Emmanuel and Kanwal, Maxinder S and Maharaj, Tegan and Fischer, Asja and Courville, Aaron and Bengio, Yoshua and others},
    booktitle = {International conference on machine learning},
    pages = {233--242},
    year = {2017},
    organization = {PMLR}
}

@article{zhong2024iclr,
    title = {ICLR: Instance credibility-based label refinement for label noisy person re-identification},
    author = {Zhong, Xian and Han, Xiyu and Jia, Xuemei and Huang, Wenxin and Liu, Wenxuan and Su, Shuaipeng and Yu, Xiaohan and Ye, Mang},
    journal = {Pattern Recognition},
    volume = {148},
    pages = {110168},
    year = {2024},
    publisher = {Elsevier}
}

@article{liu2024two,
    title = {A two-stage noise-tolerant paradigm for label corrupted person re-identification},
    author = {Liu, Min and Wang, Fei and Wang, Xueping and Wang, Yaonan and Roy-Chowdhury, Amit K},
    journal = {IEEE Transactions on Pattern Analysis and Machine Intelligence},
    volume = {46},
    number = {7},
    pages = {4944--4956},
    year = {2024},
    publisher = {IEEE}
}

@inproceedings{qin2024noisy,
    title = {Noisy-correspondence learning for text-to-image person re-identification},
    author = {Qin, Yang and Chen, Yingke and Peng, Dezhong and Peng, Xi and Zhou, Joey Tianyi and Hu, Peng},
    booktitle = {Proceedings of the IEEE/CVF Conference on Computer Vision and Pattern Recognition},
    pages = {27197--27206},
    year = {2024}
}

@article{yang2024robust,
    title = {Robust object re-identification with coupled noisy labels},
    author = {Yang, Mouxing and Huang, Zhenyu and Peng, Xi},
    journal = {International Journal of Computer Vision},
    volume = {132},
    number = {7},
    pages = {2511--2529},
    year = {2024},
    publisher = {Springer}
}

@InProceedings{Wang_2020_CVPR,
    author = {Wang, Guan'an and Yang, Shuo and Liu, Huanyu and Wang, Zhicheng and Yang, Yang and Wang, Shuliang and Yu, Gang and Zhou, Erjin and Sun, Jian},
    title = {High-Order Information Matters: Learning Relation and Topology for Occluded Person Re-Identification},
    booktitle = {Proceedings of the IEEE/CVF Conference on Computer Vision and Pattern Recognition (CVPR)},
    month = {June},
    year = {2020}
}

@inproceedings{chen2021occlude,
    title = {Occlude them all: Occlusion-aware attention network for occluded person re-id},
    author = {Chen, Peixian and Liu, Wenfeng and Dai, Pingyang and Liu, Jianzhuang and Ye, Qixiang and Xu, Mingliang and Chen, Qi’an and Ji, Rongrong},
    booktitle = {Proceedings of the IEEE/CVF international conference on computer vision},
    pages = {11833--11842},
    year = {2021}
}

@inproceedings{jiang2021exploring,
    title = {Exploring the quality of gan generated images for person re-identification},
    author = {Jiang, Yiqi and Chen, Weihua and Sun, Xiuyu and Shi, Xiaoyu and Wang, Fan and Li, Hao},
    booktitle = {Proceedings of the 29th ACM International Conference on Multimedia},
    pages = {4146--4155},
    year = {2021}
}

@inproceedings{yao2020simulating,
    title = {Simulating content consistent vehicle datasets with attribute descent},
    author = {Yao, Yue and Zheng, Liang and Yang, Xiaodong and Naphade, Milind and Gedeon, Tom},
    booktitle = {European Conference on computer vision},
    pages = {775--791},
    year = {2020},
    organization = {Springer}
}

@inproceedings{he2020multi,
  author    = {He, Shuting and Luo, Hao and Chen, Weihua and Zhang, Miao and Zhang, Yuqi and Wang, Fan and Li, Hao and Jiang, Wei},
  title     = {Multi-Domain Learning and Identity Mining for Vehicle Re-Identification},
  booktitle = {Proceedings of the IEEE/CVF Conference on Computer Vision and Pattern Recognition Workshops (CVPRW)},
  pages     = {2485--2493},
  year      = {2020},
  doi       = {10.1109/CVPRW50498.2020.00299}
}

@inproceedings{zhang2019self,
    title = {Self-training with progressive augmentation for unsupervised cross-domain person re-identification},
    author = {Zhang, Xinyu and Cao, Jiewei and Shen, Chunhua and You, Mingyu},
    booktitle = {Proceedings of the IEEE/CVF International Conference on Computer Vision},
    pages = {8222--8231},
    year = {2019}
}

@inproceedings{yang2020asymmetric,
    title = {Asymmetric co-teaching for unsupervised cross-domain person re-identification},
    author = {Yang, Fengxiang and Li, Ke and Zhong, Zhun and Luo, Zhiming and Sun, Xing and Cheng, Hao and Guo, Xiaowei and Huang, Feiyue and Ji, Rongrong and Li, Shaozi},
    booktitle = {Proceedings of the AAAI conference on artificial intelligence},
    volume = {34},
    pages = {12597--12604},
    year = {2020}
}

@inproceedings{zhao2020unsupervised,
    title = {Unsupervised domain adaptation with noise resistible mutual-training for person re-identification},
    author = {Zhao, Fang and Liao, Shengcai and Xie, Guo-Sen and Zhao, Jian and Zhang, Kaihao and Shao, Ling},
    booktitle = {European conference on computer vision},
    pages = {526--544},
    year = {2020},
    organization = {Springer}
}

@article{SUN2024105244,
    title = {Pro-ReID: Producing reliable pseudo labels for unsupervised person re-identification},
    journal = {Image and Vision Computing},
    volume = {150},
    pages = {105244},
    year = {2024},
    doi = {https://doi.org/10.1016/j.imavis.2024.105244},
    author = {Haiming Sun and Shiwei Ma},
}

@article{dutta2026latent,
    title = {Latent Diffusion-Guided Feature Inpainting for Occluded Person Re-Identification With Hybrid Re-Ranking},
    author = {Dutta, Pratyay and Bhanu, Bir},
    journal = {IEEE Transactions on Circuits and Systems for Video Technology},
    year = {2026},
    publisher = {IEEE}
}

@inproceedings{Wang_2022_CVPR,
  author    = {Wang, Zhikang and Zhu, Feng and Tang, Shixiang and Zhao, Rui and He, Lihuo and Song, Jiangning},
  title     = {Feature Erasing and Diffusion Network for Occluded Person Re-Identification},
  booktitle = {Proceedings of the IEEE/CVF Conference on Computer Vision and Pattern Recognition (CVPR)},
  pages     = {4744--4753},
  year      = {2022},
  doi       = {10.1109/CVPR52688.2022.00471}
}

@InProceedings{Gao_2020_CVPR,
    author = {Gao, Shang and Wang, Jingya and Lu, Huchuan and Liu, Zimo},
    title = {Pose-Guided Visible Part Matching for Occluded Person ReID},
    booktitle = {Proceedings of the IEEE/CVF Conference on Computer Vision and Pattern Recognition (CVPR)},
    month = {June},
    year = {2020}
}

@InProceedings{Sun_2019_CVPR,
    author = {Sun, Yifan and Xu, Qin and Li, Yali and Zhang, Chi and Li, Yikang and Wang, Shengjin and Sun, Jian},
    title = {Perceive Where to Focus: Learning Visibility-Aware Part-Level Features for Partial Person Re-Identification},
    booktitle = {Proceedings of the IEEE/CVF Conference on Computer Vision and Pattern Recognition (CVPR)},
    month = {June},
    year = {2019}
}

@InProceedings{Somers_2023_WACV,
    author = {Somers, Vladimir and De Vleeschouwer, Christophe and Alahi, Alexandre},
    title = {Body Part-Based Representation Learning for Occluded Person Re-Identification},
    booktitle = {Proceedings of the IEEE/CVF Winter Conference on Applications of Computer Vision (WACV)},
    month = {January},
    year = {2023},
    pages = {1613-1623}
}

@InProceedings{Miao_2019_ICCV,
    author = {Miao, Jiaxu and Wu, Yu and Liu, Ping and Ding, Yuhang and Yang, Yi},
    title = {Pose-Guided Feature Alignment for Occluded Person Re-Identification},
    booktitle = {Proceedings of the IEEE/CVF International Conference on Computer Vision (ICCV)},
    month = {October},
    year = {2019}
}

@misc{zhao2025hunyuan3d,
    title = {Hunyuan3D 2.0: Scaling Diffusion Models for High Resolution Textured 3D Assets Generation},
    author = {Zibo Zhao and Zeqiang Lai and Qingxiang Lin and Yunfei Zhao and Haolin Liu and Shuhui Yang and Yifei Feng and Mingxin Yang and Sheng Zhang and Xianghui Yang and Huiwen Shi and Sicong Liu and Junta Wu and Yihang Lian and Fan Yang and Ruining Tang and Zebin He and Xinzhou Wang and Jian Liu and Xuhui Zuo and Zhuo Chen and Biwen Lei and Haohan Weng and Jing Xu and Yiling Zhu and Xinhai Liu and Lixin Xu and Changrong Hu and Shaoxiong Yang and Song Zhang and Yang Liu and Tianyu Huang and Lifu Wang and Jihong Zhang and Meng Chen and Liang Dong and Yiwen Jia and Yulin Cai and Jiaao Yu and Yixuan Tang and Hao Zhang and Zheng Ye and Peng He and Runzhou Wu and Chao Zhang and Yonghao Tan and Jie Xiao and Yangyu Tao and Jianchen Zhu and Jinbao Xue and Kai Liu and Chongqing Zhao and Xinming Wu and Zhichao Hu and Lei Qin and Jianbing Peng and Zhan Li and Minghui Chen and Xipeng Zhang and Lin Niu and Paige Wang and Yingkai Wang and Haozhao Kuang and Zhongyi Fan and Xu Zheng and Weihao Zhuang and YingPing He and Tian Liu and Yong Yang and Di Wang and Yuhong Liu and Jie Jiang and Jingwei Huang and Chunchao Guo},
    year = {2025},
    eprint = {2501.12202},
    archivePrefix = {arXiv},
    primaryClass = {cs.CV},
}

@inproceedings{luo2019bag,
  author    = {Luo, Hao and Gu, Youzhi and Liao, Xingyu and Lai, Shenqi and Jiang, Wei},
  title     = {Bag of Tricks and a Strong Baseline for Deep Person Re-Identification},
  booktitle = {Proceedings of the IEEE/CVF Conference on Computer Vision and Pattern Recognition Workshops (CVPRW)},
  pages     = {1487--1495},
  year      = {2019},
  doi       = {10.1109/CVPRW.2019.00190}
}

@inproceedings{dosovitskiy2021image,
    title = {An Image is Worth 16x16 Words: Transformers for Image Recognition at Scale},
    author = {Dosovitskiy, Alexey and Beyer, Lucas and Kolesnikov, Alexander and Weissenborn, Dirk and Zhai, Xiaohua and Unterthiner, Thomas and Dehghani, Mostafa and Minderer, Matthias and Heigold, Georg and Gelly, Sylvain and Uszkoreit, Jakob and Houlsby, Neil},
    booktitle = ICLR,
    year = {2021}
}

@inproceedings{he2023fastreid,
    title = {Fastreid: A pytorch toolbox for general instance re-identification},
    author = {He, Lingxiao and Liao, Xingyu and Liu, Wu and Liu, Xinchen and Cheng, Peng and Mei, Tao},
    booktitle = ACMMM,
    pages = {9664--9667},
    year = {2023}
}

@article{ye2021deep,
  author    = {Ye, Mang and Shen, Jianbing and Lin, Gaojie and Xiang, Tao and Shao, Ling and Hoi, Steven C. H.},
  title     = {Deep Learning for Person Re-Identification: A Survey and Outlook},
  journal   = {IEEE Transactions on Pattern Analysis and Machine Intelligence},
  volume    = {44},
  number    = {6},
  pages     = {2872--2893},
  year      = {2022},
  doi       = {10.1109/TPAMI.2021.3054775}
}

@inproceedings{he2021transreid,
  author    = {He, Shuting and Luo, Hao and Wang, Pichao and Wang, Fan and Li, Hao and Jiang, Wei},
  title     = {TransReID: Transformer-based Object Re-Identification},
  booktitle = {Proceedings of the IEEE/CVF International Conference on Computer Vision (ICCV)},
  pages     = {14993--15002},
  year      = {2021},
  doi       = {10.1109/ICCV48922.2021.01474}
}

@inproceedings{zhu2022pass,
    title = {Pass: Part-aware self-supervised pre-training for person re-identification},
    author = {Zhu, Kuan and Guo, Haiyun and Yan, Tianyi and Zhu, Yousong and Wang, Jinqiao and Tang, Ming},
    booktitle = ECCV,
    pages = {198--214},
    year = {2022}
}

@inproceedings{chen2022rotation,
    title = {Rotation invariant transformer for recognizing object in uavs},
    author = {Chen, Shuoyi and Ye, Mang and Du, Bo},
    booktitle = ACMMM,
    pages = {2565--2574},
    year = {2022}
}

@inproceedings{li2023dc,
    title = {Dc-former: Diverse and compact transformer for person re-identification},
    author = {Li, Wen and Zou, Cheng and Wang, Meng and Xu, Furong and Zhao, Jianan and Zheng, Ruobing and Cheng, Yuan and Chu, Wei},
    booktitle = AAAI,
    volume = {37},
    pages = {1415--1423},
    year = {2023}
}

@misc{ridnik2021imagenet,
    title = {ImageNet-21K Pretraining for the Masses},
    author = {Tal Ridnik and Emanuel Ben-Baruch and Asaf Noy and Lihi Zelnik-Manor},
    year = {2021},
    eprint = {2104.10972},
    archivePrefix = {arXiv},
    primaryClass = {cs.CV},
}

@misc{grolleau2026moomultivieworientedobservations,
    title = {MOO: A Multi-view Oriented Observations Dataset for Viewpoint Analysis in Cattle Re-Identification},
    author = {Grolleau, William and Chaouch, Achraf and Sabourin, Astrid and Lapouge, Guillaume and Achard, Catherine},
    year = {2026},
    eprint = {2603.04314},
    archivePrefix = {arXiv},
    primaryClass = {cs.CV}
}

@inproceedings{chung2024styleid,
    title = {Style Injection in Diffusion: A Training-free Approach for Adapting Large-scale Diffusion Models for Style Transfer},
    author = {Chung, Jiwoo and Hyun, Sangeek and Heo, Jae-Pil},
    booktitle = {Proceedings of the IEEE/CVF Conference on Computer Vision and Pattern Recognition (CVPR)},
    pages = {8795--8805},
    year = {2024}
}

    \newpage
    \appendix

    \section{Supplementary Material}

    \subsection{Visibility Map Analysis}

    \Cref{tab:ablation_visibility_mode} compares our mesh-derived visibility map against alternatives that ignore which regions were actually visible.
    The One maps apply uniform attention supervision, while the Foreground map restricts attention to the rendered subject without distinguishing visible from hallucinated regions.
    Both reach roughly the same accuracy as each other on AG-ReID Aerial and are well below our visibility map with a gap of $+1.5$ to $+1.7$ mAP pp on Aerial training.
    This shows that the gain from VAAS does not come from subject masking but from the per-patch reliability information that only the mesh can supply.
    Conversely, directing attention to the hallucinated regions falls below the model trained without attention supervision (Index 4 of \Cref{tab:comp_components}) in every column, confirming that visible regions carry more identity signal than hallucinated ones.

    \Cref{tab:ablation_masked} shows that hallucinated regions should nevertheless not be removed from the input.
    Hard-masking hallucinated pixels in the rendered views lowers accuracy in every column, since these regions still provide object context such as silhouette and pose.
    VAAS therefore guides the identity token toward visible patches without excluding hallucinated content from the representation.

    \subsection{Low-Visibility Samples}

    \Cref{tab:ablation_hallu_ratio} retains only the novel views whose hallucinated ratio is below a threshold.
    Accuracy increases as more hallucinated views are retained, and keeping all views is best in every column.
    Low-visibility views come from the largest viewpoint shifts and are relevant for large viewpoint changes despite their lower share of visible pixels.
    For this reason, CVCF targets geometrically incorrect meshes rather than low-visibility views.

    \subsection{Calibration Noise}

    The visibility map depends on the calibrated source pose $\Pi_i$, for which no ground truth is available.
    We therefore measure the downstream effect of calibration errors by adding noise of up to $\pm n^\circ$ in azimuth and elevation to the fitted pose (\Cref{tab:ablation_calib}).
    A $\pm 5^\circ$ perturbation leaves accuracy almost unchanged, while larger perturbations degrade it progressively.
    For scale, the fitted rotations average $21.6^\circ / 27.0^\circ / 0.5^\circ$ (azimuth / elevation / roll) on AG-ReID, $19.3^\circ / 20.7^\circ / 0.3^\circ$ on AG-ReID.v2, and $23.0^\circ / 36.9^\circ / 0.8^\circ$ on MOO.
    The visibility map is therefore robust to small calibration errors, and CVCF removes the meshes whose calibration fails.

    \begin{table}[tbp]
        \caption{
            \textbf{Hallucinated Pixel Masking on AG-ReID.}
            Hallucinated pixels of the rendered views are masked out of the input.
            mAP and Rank-1 (\%) on Aerial $\rightarrow$ Ground and Ground $\rightarrow$ Aerial.
            Best in \textbf{bold}.
        }
        \label{tab:ablation_masked}
        \centering
        \scriptsize

        \begin{tabular}{l cc cc cc cc}
            \toprule
            & \multicolumn{4}{c}{Aerial Training} & \multicolumn{4}{c}{Ground Training} \\
            \cmidrule(lr){2-5} \cmidrule(lr){6-9}
            \textbf{Variant}
            & \multicolumn{2}{c}{$A \rightarrow G$}
            & \multicolumn{2}{c}{$G \rightarrow A$}
            & \multicolumn{2}{c}{$A \rightarrow G$}
            & \multicolumn{2}{c}{$G \rightarrow A$} \\
            \cmidrule(lr){2-3} \cmidrule(lr){4-5} \cmidrule(lr){6-7} \cmidrule(lr){8-9}
            & mAP           & R-1           & mAP           & R-1           & mAP           & R-1           & mAP           & R-1           \\
            \midrule
            Visibility map (ours)      & \textbf{52.1} & \textbf{66.1} & \textbf{53.2} & \textbf{62.6} & \textbf{56.0} & \textbf{67.0} & \textbf{57.9} & \textbf{68.6} \\
            Hallucinated pixels masked & 50.3          & 63.1          & 46.5          & 52.4          & 54.3          & 63.0          & 55.3          & 66.0          \\
            \bottomrule
        \end{tabular}
    \end{table}

    \begin{table}[tbp]
        \caption{
            \textbf{Hallucinated Fraction Ablation on AG-ReID.}
            Only novel views with less than the given fraction of hallucinated pixels are retained.
            mAP and Rank-1 (\%) on Aerial $\rightarrow$ Ground and Ground $\rightarrow$ Aerial.
            Best in \textbf{bold}, second-best \underline{underlined}.
        }
        \label{tab:ablation_hallu_ratio}
        \centering
        \scriptsize

        \begin{tabular}{l cc cc cc cc}
            \toprule
            & \multicolumn{4}{c}{Aerial Training} & \multicolumn{4}{c}{Ground Training} \\
            \cmidrule(lr){2-5} \cmidrule(lr){6-9}
            \textbf{Variant}
            & \multicolumn{2}{c}{$A \rightarrow G$}
            & \multicolumn{2}{c}{$G \rightarrow A$}
            & \multicolumn{2}{c}{$A \rightarrow G$}
            & \multicolumn{2}{c}{$G \rightarrow A$} \\
            \cmidrule(lr){2-3} \cmidrule(lr){4-5} \cmidrule(lr){6-7} \cmidrule(lr){8-9}
            & mAP              & R-1              & mAP              & R-1              & mAP              & R-1              & mAP              & R-1              \\
            \midrule
            $<10\%$ hallucinated & 47.6             & 62.9             & 48.4             & 57.6             & 51.3             & 62.3             & 53.5             & 64.3             \\
            $<30\%$ hallucinated & 49.1             & 63.6             & 51.2             & 59.6             & 52.8             & 64.6             & 55.3             & 66.3             \\
            $<50\%$ hallucinated & 50.9             & 64.7             & 51.7             & 60.9             & 54.1             & \underline{66.4} & 56.9             & 67.2             \\
            $<70\%$ hallucinated & \underline{51.5} & \underline{65.8} & \underline{52.5} & \underline{61.3} & \underline{55.3} & 65.7 & \underline{57.6} & \underline{67.7} \\
            All views (ours)     & \textbf{52.1}    & \textbf{66.1}    & \textbf{53.2}    & \textbf{62.6}    & \textbf{56.0} & \textbf{67.0} & \textbf{57.9} & \textbf{68.6} \\
            \bottomrule
        \end{tabular}
    \end{table}

    \begin{table}[tbp]
        \caption{
            \textbf{Calibration Noise Ablation on AG-ReID.}
            Noise of up to $\pm n^\circ$ in azimuth and elevation is added to the fitted source pose.
            mAP and Rank-1 (\%) on Aerial $\rightarrow$ Ground and Ground $\rightarrow$ Aerial.
            Best in \textbf{bold}, second-best \underline{underlined}.
        }
        \label{tab:ablation_calib}
        \centering
        \scriptsize

        \begin{tabular}{l cc cc cc cc}
            \toprule
            & \multicolumn{4}{c}{Aerial Training} & \multicolumn{4}{c}{Ground Training} \\
            \cmidrule(lr){2-5} \cmidrule(lr){6-9}
            \textbf{Variant}
            & \multicolumn{2}{c}{$A \rightarrow G$}
            & \multicolumn{2}{c}{$G \rightarrow A$}
            & \multicolumn{2}{c}{$A \rightarrow G$}
            & \multicolumn{2}{c}{$G \rightarrow A$} \\
            \cmidrule(lr){2-3} \cmidrule(lr){4-5} \cmidrule(lr){6-7} \cmidrule(lr){8-9}
            & mAP              & R-1              & mAP              & R-1              & mAP              & R-1              & mAP              & R-1              \\
            \midrule
            No noise (ours) & \textbf{52.1}    & \textbf{66.1}    & \textbf{53.2}    & \textbf{62.6}    & \textbf{56.0} & \textbf{67.0} & \textbf{57.9} & \textbf{68.6} \\
            $\pm 5^\circ$   & \underline{52.0} & \textbf{66.1}    & \underline{52.9} & \underline{62.4} & \underline{55.8} & \underline{66.9} & \underline{57.8} & \underline{68.4} \\
            $\pm 10^\circ$  & 51.2             & \underline{65.8} & 52.6             & 61.9             & 53.4             & 63.9             & 56.8             & 67.7             \\
            $\pm 20^\circ$  & 49.7             & 63.9             & 50.6             & 60.1             & 52.1             & 62.0             & 54.2             & 66.8             \\
            $\pm 30^\circ$  & 48.1             & 60.4             & 47.4             & 57.9             & 50.4             & 59.2             & 52.2             & 62.1             \\
            \bottomrule
        \end{tabular}
    \end{table}

    \begin{table}[tbp]
        \caption{
            \textbf{Visibility Mask Source Ablation on AG-ReID.}
            mAP and Rank-1 (\%) on Aerial $\rightarrow$ Ground and Ground $\rightarrow$ Aerial.
            Best in \textbf{bold}, second-best \underline{underlined}.
        }
        \label{tab:ablation_visibility_mode}
        \centering

        \scriptsize

        \begin{tabular}{l cc cc cc cc}
            \toprule
            & \multicolumn{4}{c}{Aerial Training} & \multicolumn{4}{c}{Ground Training} \\
            \cmidrule(lr){2-5} \cmidrule(lr){6-9}
            \textbf{Variant}
            & \multicolumn{2}{c}{$A \rightarrow G$}
            & \multicolumn{2}{c}{$G \rightarrow A$}
            & \multicolumn{2}{c}{$A \rightarrow G$}
            & \multicolumn{2}{c}{$G \rightarrow A$} \\
            \cmidrule(lr){2-3} \cmidrule(lr){4-5} \cmidrule(lr){6-7} \cmidrule(lr){8-9}
            & mAP              & R-1              & mAP              & R-1              & mAP              & R-1              & mAP              & R-1              \\
            \midrule
            One map               & 50.5             & 64.0             & 51.5             & \underline{59.4} & \underline{55.0} & \underline{66.2} & 54.4             & \underline{65.3} \\
            Foreground map        & \underline{50.6} & \underline{65.1} & \underline{51.6} & 59.2             & 54.8             & 65.3 & 53.5 & 64.9 \\
            Hallucinated map      & 48.2             & 62.5             & 50.3             & 57.8             & 52.8             & 63.3             & \underline{55.3} & \underline{65.3} \\
            Visibility map (ours) & \textbf{52.1}    & \textbf{66.1}    & \textbf{53.2}    & \textbf{62.6}    & \textbf{56.0} & \textbf{67.0} & \textbf{57.9} & \textbf{68.6} \\
            \bottomrule
        \end{tabular}
    \end{table}

    \subsection{Number of supervised layers}

    \Cref{tab:ablation_n_layers} tries different values of $L$, the number of transformer layers on which the CLS-attention loss is applied.
    Supervising only the last layer ($L=1$) is best across all eight columns, with $52.1$ mAP and $66.1$ R-1 on Aerial $A\rightarrow G$.
    Extending the supervision to more layers degrades performance until it stabilizes at $L=4$ and beyond.
    The visibility map specifies where the final identity representation should attend, but constraining earlier layers removes their ability to build the global context that the last-layer attention then exploits.

    \subsection{Triplet loss weight}

    \Cref{tab:ablation_lambda_a} sweeps the weight $\lambda_{\text{tri}}$ of the visibility-weighted triplet loss.
    Setting $\lambda_{\text{tri}}=0.25$ underuses the term and yields the weakest results ($51.1$ mAP and $59.6$ R-1 on Aerial $A\rightarrow G$), while $\lambda_{\text{tri}}\in\{0.5, 0.75, 1.0\}$ all reach comparable accuracy.
    This shows that the visibility-weighted triplet loss is a key component of our method, but the exact weight is not critical as long as it is sufficiently high to influence training.

    \subsection{CLS-attention loss weight}
    \Cref{tab:ablation_lambda_d} sweeps the weight $\lambda_{\text{vis}}$ of the CLS-attention loss with $\lambda_{\text{tri}}=1$ fixed.
    Again, setting $\lambda_{\text{vis}}=0.25$ does not provide enough signal ($51.3$ mAP on Aerial $A \rightarrow G$) and higher values yield comparable performance.
    $\lambda_{\text{vis}}=1$ is best overall, and we adopt it in our experiments.

    \begin{table}[tbp]
        \caption{
            \textbf{Number of Supervised Attention Layers Ablation on AG-ReID.}
            mAP and Rank-1 (\%) on Aerial $\rightarrow$ Ground and Ground $\rightarrow$ Aerial.
            Best in \textbf{bold}, second-best \underline{underlined}.
        }
        \label{tab:ablation_n_layers}
        \centering

        \scriptsize

        \begin{tabular}{l cc cc cc cc}
            \toprule
            & \multicolumn{4}{c}{Aerial Training} & \multicolumn{4}{c}{Ground Training} \\
            \cmidrule(lr){2-5} \cmidrule(lr){6-9}
            \textbf{Variant}
            & \multicolumn{2}{c}{$A \rightarrow G$}
            & \multicolumn{2}{c}{$G \rightarrow A$}
            & \multicolumn{2}{c}{$A \rightarrow G$}
            & \multicolumn{2}{c}{$G \rightarrow A$} \\
            \cmidrule(lr){2-3} \cmidrule(lr){4-5} \cmidrule(lr){6-7} \cmidrule(lr){8-9}
            & mAP              & R-1              & mAP              & R-1              & mAP              & R-1              & mAP              & R-1              \\
            \midrule
            $L=2$        & \underline{51.2} & \underline{64.4} & \underline{52.3} & \underline{61.4} & \underline{55.2} & \underline{65.4}             & \underline{56.4} & 66.1 \\
            $L=4$        & 50.1             & 63.8             & 51.2             & 59.2             & 54.0             & 64.0             & 55.7             & 66.0             \\
            $L=8$        & 50.0             & 63.8             & 51.2             & 59.2             & 53.8             & 64.8             & 55.6             & \underline{66.4} \\
            $L=12$       & 50.2             & 63.7             & 51.0             & 59.5             & 53.6             & 64.7             & 55.7             & 65.7             \\
            $L=1$ (ours) & \textbf{52.1}    & \textbf{66.1}    & \textbf{53.2}    & \textbf{62.6}    & \textbf{56.0}    & \textbf{67.0}    & \textbf{57.9}    & \textbf{68.6}    \\
            \bottomrule
        \end{tabular}
    \end{table}

    \begin{table}[tbp]
        \caption{
            \textbf{Triplet Loss Weight $\lambda_{\text{tri}}$ Ablation on AG-ReID.}
            mAP and Rank-1 (\%) on Aerial $\rightarrow$ Ground and Ground $\rightarrow$ Aerial.
            Best in \textbf{bold}, second-best \underline{underlined}.
        }
        \label{tab:ablation_lambda_a}
        \centering

        \scriptsize

        \begin{tabular}{l cc cc cc cc}
            \toprule
            & \multicolumn{4}{c}{Aerial Training} & \multicolumn{4}{c}{Ground Training} \\
            \cmidrule(lr){2-5} \cmidrule(lr){6-9}
            \textbf{Variant}
            & \multicolumn{2}{c}{$A \rightarrow G$}
            & \multicolumn{2}{c}{$G \rightarrow A$}
            & \multicolumn{2}{c}{$A \rightarrow G$}
            & \multicolumn{2}{c}{$G \rightarrow A$} \\
            \cmidrule(lr){2-3} \cmidrule(lr){4-5} \cmidrule(lr){6-7} \cmidrule(lr){8-9}
            & mAP              & R-1              & mAP              & R-1              & mAP              & R-1              & mAP              & R-1              \\
            \midrule
            $\lambda_{\text{tri}}{=}0.25$        & 51.1             & 59.6             & 49.7             & 59.4             & 54.4             & 65.7             & 56.6             & 66.4             \\
            $\lambda_{\text{tri}}{=}0.50$        & 51.5             & \underline{65.8} & \underline{52.9} & \underline{61.3} & \underline{55.9}             & \underline{66.9}             & 55.3 & 66.5 \\
            $\lambda_{\text{tri}}{=}0.75$        & \underline{51.9} & 64.8             & \underline{52.9} & \underline{61.3}             & 55.8 & \underline{66.9}    & \underline{57.8}             & \textbf{68.9} \\
            $\lambda_{\text{tri}}{=}1.00$ (ours) & \textbf{52.1}    & \textbf{66.1}    & \textbf{53.2}    & \textbf{62.6}    & \textbf{56.0}    & \textbf{67.0} & \textbf{57.9}    & \underline{68.6}    \\
            \bottomrule
        \end{tabular}
    \end{table}

    \begin{table}[tbp]
        \caption{
            \textbf{CLS-Attention Loss Weight $\lambda_{\text{vis}}$ Ablation on AG-ReID.}
            mAP and Rank-1 (\%) on Aerial $\rightarrow$ Ground and Ground $\rightarrow$ Aerial.
            Best in \textbf{bold}, second-best \underline{underlined}.
        }
        \label{tab:ablation_lambda_d}
        \centering

        \scriptsize

        \begin{tabular}{l cc cc cc cc}
            \toprule
            & \multicolumn{4}{c}{Aerial Training} & \multicolumn{4}{c}{Ground Training} \\
            \cmidrule(lr){2-5} \cmidrule(lr){6-9}
            \textbf{Variant}
            & \multicolumn{2}{c}{$A \rightarrow G$}
            & \multicolumn{2}{c}{$G \rightarrow A$}
            & \multicolumn{2}{c}{$A \rightarrow G$}
            & \multicolumn{2}{c}{$G \rightarrow A$} \\
            \cmidrule(lr){2-3} \cmidrule(lr){4-5} \cmidrule(lr){6-7} \cmidrule(lr){8-9}
            & mAP              & R-1              & mAP              & R-1              & mAP              & R-1              & mAP              & R-1              \\
            \midrule
            $\lambda_{\text{vis}}{=}0.25$        & 51.3             & 64.8             & 51.9             & 59.5             & 54.2             & 65.2             & 56.5             & 65.9             \\
            $\lambda_{\text{vis}}{=}0.50$        & 51.8             & \underline{65.7} & 52.0             & 60.6             & \underline{55.9} & \underline{66.8} & \underline{57.5} & \underline{68.5} \\
            $\lambda_{\text{vis}}{=}0.75$        & \underline{51.9} & 65.5             & \underline{52.7} & \underline{61.1}             & 55.6             & 66.1 & 57.4             & \underline{68.5}             \\
            $\lambda_{\text{vis}}{=}1.00$ (ours) & \textbf{52.1}    & \textbf{66.1}    & \textbf{53.2}    & \textbf{62.6}    & \textbf{56.0}    & \textbf{67.0}    & \textbf{57.9}    & \textbf{68.6}    \\
            \bottomrule
        \end{tabular}
    \end{table}

    \begin{table}[tbp]
        \caption{
            \textbf{Filtering Ratio Ablation on AG-ReID.}
            Percentage of filtered novel meshes. mAP and Rank-1 (\%) on Aerial $\rightarrow$ Ground and Ground $\rightarrow$ Aerial.
            Best in \textbf{bold}, second-best \underline{underlined}.
        }
        \label{tab:ablation_filter_tau}
        \centering

        \scriptsize

        \begin{tabular}{l cc cc cc cc}
            \toprule
            & \multicolumn{4}{c}{Aerial Training} & \multicolumn{4}{c}{Ground Training} \\
            \cmidrule(lr){2-5} \cmidrule(lr){6-9}
            \textbf{Variant}
            & \multicolumn{2}{c}{$A \rightarrow G$}
            & \multicolumn{2}{c}{$G \rightarrow A$}
            & \multicolumn{2}{c}{$A \rightarrow G$}
            & \multicolumn{2}{c}{$G \rightarrow A$} \\
            \cmidrule(lr){2-3} \cmidrule(lr){4-5} \cmidrule(lr){6-7} \cmidrule(lr){8-9}
            & mAP              & R-1              & mAP              & R-1              & mAP              & R-1              & mAP              & R-1              \\
            \midrule
            $0.10$        & 49.2             & 62.1             & 45.9             & 53.3             & 53.1             & 62.5             & 56.2             & 68.1             \\
            $0.20$        & 50.6             & 62.3             & 51.5             & 59.5             & 55.3             & 66.0             & 57.1             & \textbf{68.7}    \\
            $0.30$        & 51.7             & 63.9             & 52.9             & 61.8             & 55.3             & 65.3             & 57.3             & 67.2             \\
            $0.40$        & \underline{51.8} & \underline{66.0} & \underline{53.0} & \underline{62.4} & 55.3             & 65.9             & 57.6             & 67.8             \\
            $0.50$ (ours) & \textbf{52.1}    & \textbf{66.1}    & \textbf{53.2}    & \textbf{62.6}    & \textbf{56.0}    & \textbf{67.0} & \textbf{57.9}    & \underline{68.6}    \\
            $0.60$        & 51.1             & 65.4             & \underline{53.0} & 62.0             & 55.3             & 66.2             & \underline{57.8} & 68.5             \\
            $0.70$        & 51.2             & \underline{66.0} & \underline{53.0} & 61.1             & \underline{55.5} & \underline{66.9}             & 57.7 & 68.1 \\
            $0.80$        & 50.8             & 65.5             & 52.8             & 60.9             & 54.6             & 65.5             & 57.3             & 68.3             \\
            $0.90$        & 50.1             & 64.2             & 52.3             & 61.0             & 53.2             & 64.2             & 55.5             & 65.7             \\
            \bottomrule
        \end{tabular}
    \end{table}

    \subsection{Filtering ratio}

    \Cref{tab:ablation_filter_tau} reports the effect of the filtering percentage, which controls the quantity of filtered novel meshes discarded.
    At low values ($\le 0.3$) filtering is too permissive and low-quality synthesized samples with severe artifacts or poor geometric fidelity leak into training and drag performance down across both protocols, with $0.1$ falling to 49.2 mAP on $A{\rightarrow}G$.
    At high values ($\ge 0.7$) filtering becomes overly aggressive and prunes informative views along with the noisy ones, eroding the diversity needed to bridge the aerial-to-ground viewpoint gap.
    We chose to filter the bottom $50\%$ of samples, which is a good compromise between quality and diversity.
    However, the ideal threshold depends on the complexity of the dataset and on the robustness of the 3D reconstruction method, which may require tuning for different settings.

    \subsection{Computational Cost}
    \label{subsec:computational-cost}
    The initial 3D lifting of source images is the most computationally intensive task of our pipeline, but it amortizes the overall cost by enabling rendering of multiple novel views per mesh.
    Crucially, this cost is incurred only at training time.
    At inference, 3D-VITAL uses the same ViT backbone as the baseline, since our components only modify the training data and loss function.

    Specifically, for AG-ReID, the 3D reconstruction of 10{,}000 source images (all successfully segmented images of the training set) requires 24 hours on 8 A100 GPUs.
    Rendering 100{,}000 novel views and their associated visibility maps (10 views for each of the 10{,}000 meshes) takes only 8 hours on one P5000 GPU.
    Finally, the StyleID step requires 8 hours on 8 A100 GPUs.
    The final ReID training with 3D-VITAL is equivalent to the ViT baseline, as our components are not computationally intensive and do not require additional forward passes at inference time.
    Including the remaining generation stages, the pipeline requires 272 A100 and 8 P5000 GPU-hours for the entire AG-ReID training set.
    This cost is a one-time investment that is independent of downstream ReID training.

    For rough context, diffusion-based augmentation methods such as Diff-Person~\cite{niu2025synthesizing} require per-identity fine-tuning of a diffusion model and report three days on 8 RTX 4090 GPUs (576 GPU-hours) for 5{,}000 identities, which must be repeated whenever the identity set changes.
    They do not report the cost of filtering their generated samples within the paper.
    A100 and RTX 4090 GPUs are not directly comparable in compute throughput, so these numbers should be read as orders of magnitude rather than a precise ratio.

\end{document}